\documentclass[11pt]{article}

\usepackage[preprint]{acl}

\usepackage{times}
\usepackage{latexsym}

\usepackage[T1]{fontenc}
\usepackage[utf8]{inputenc}

\usepackage{microtype}

\usepackage{graphicx}
\usepackage{bbm}
\usepackage{amsmath}
\usepackage{subcaption}
\usepackage{graphicx}

\usepackage{tabularx}
\usepackage{booktabs}
\usepackage{threeparttable}
\usepackage[table]{xcolor}
\usepackage{pifont}

\newcommand{\graytext}[1]{\textcolor{gray}{#1}}
\usepackage{multirow}
\usepackage{booktabs}
\usepackage{threeparttable}
\usepackage[table]{xcolor}

\usepackage[most]{tcolorbox}
\usepackage{enumitem}
\usepackage{hyperref}
\usepackage{cleveref}
\usepackage{sourcesanspro}
\usepackage[varqu,varl]{inconsolata}
\usepackage[table]{xcolor}
\definecolor{promptbg}{HTML}{F8FAFC}
\definecolor{promptborder}{HTML}{CBD5E1}
\definecolor{promptheader}{HTML}{0F172A}
\definecolor{promptlabel}{HTML}{1E293B}
\definecolor{prompttext}{HTML}{334155}
\definecolor{promptplaceholderbg}{HTML}{E2E8F0}
\definecolor{promptplaceholder}{HTML}{0F172A}

\newtcolorbox[auto counter]{promptbox}[2][]{
    enhanced,
    breakable,
    colback=promptbg,
    colframe=promptborder,
    coltitle=white,
    colbacktitle=promptheader,
    title=\textbf{Prompt~\thetcbcounter: #2},
    fonttitle=\small\sffamily,
    fontupper=\small\sffamily\color{prompttext},
    boxrule=0.7pt,
    arc=2mm,
    left=3mm,
    right=3mm,
    top=2mm,
    bottom=2mm,
    toptitle=1mm,
    bottomtitle=1mm,
    titlerule=0pt,
    before upper={
        \setlength{\parindent}{0pt}
        \setlength{\parskip}{4pt}
        \linespread{1.03}\selectfont
    },
    #1
}

\newcommand{\placeholder}[1]{%
    \tcbox[
        on line,
        colback=promptplaceholderbg,
        colframe=promptplaceholderbg,
        arc=1mm,
        boxrule=0pt,
        left=1mm,
        right=1mm,
        top=0.3mm,
        bottom=0.3mm
    ]{\texttt{\color{promptplaceholder}\{#1\}}}%
}

\crefname{tcbcounter}{Prompt}{Prompts}

\title{Context Distillation as Latent Memory Management}

\author{
 \textbf{Ziyang Zheng\textsuperscript{1,2}},
 \textbf{Zeju Li\textsuperscript{1}},
 \textbf{Xiangyu Wen\textsuperscript{1}},
 \textbf{Jianyuan Zhong\textsuperscript{1}},
\\
 \textbf{Junhua Huang\textsuperscript{2}},
 \textbf{Lei Chen\textsuperscript{2}},
 \textbf{Mingxuan Yuan\textsuperscript{2}},
 \textbf{Qiang Xu \textsuperscript{1}}
\\
\\
 \textsuperscript{1}The Chinese University of Hong Kong,
 \textsuperscript{2}Huawei Noah's Ark Lab
\\
 \small{
   \textbf{Correspondence:} \href{mailto:email@domain}{qxu@cse.cuhk.edu.hk}
 }
}

\begin{document}
\maketitle

\begin{abstract}
Context distillation compresses contextual information into model parameters, yet existing methods often ignore how multiple distilled latent memories should be stored, retrieved, and safely activated in non-oracle settings. We formulate context distillation as a latent memory management problem. We distill each context into an independent LoRA adapter, forming a modular memory bank that enables explicit memory selection. Given a query, our framework retrieves candidate memories, routes the query to the most suitable adapter, and uses a Self-Gating mechanism to decide whether latent memory should be activated. To improve efficiency, we further introduce cache sharing to reduce management overhead during inference. Experiments show that our method substantially outperforms baselines with retrieval, while Self-Gating improves robustness by deactivate unnecessary latent memories.
\end{abstract}

\section{Introduction}
\vspace{-5pt}
LLMs adapt to documents, tasks, and users primarily by placing relevant information in the context window~\citep{brown2020language,lewis2020retrieval}. 
This in-context adaptation is flexible but inherently temporary: the same information must be re-read for every query, and long prompts increase latency~\citep{tay2022efficient}, memory usage, and generation instability~\citep{zhao2021calibrate,liu2024lost}. 
These limitations raise a central question: \textit{Can contextual information be converted from temporary text into persistent model memory, and if so, how should many such memories be stored, retrieved, and safely activated?}

Context distillation (CD)~\citep{cao2025infiniteicl,wang2024templora,ye2026OPCD,zhang2026opsdl,charakorn2026doc2lora,charakorn2025text2lora} offers a natural answer to the first part of this question. 
Instead of conditioning on a long context at every inference step, CD trains a model to reproduce its context-conditioned behavior without explicitly seeing the context, thereby compressing textual context into latent memory stored in model parameters for future use.
\begin{figure}[t]
    \centering
    \includegraphics[width=0.85\linewidth]{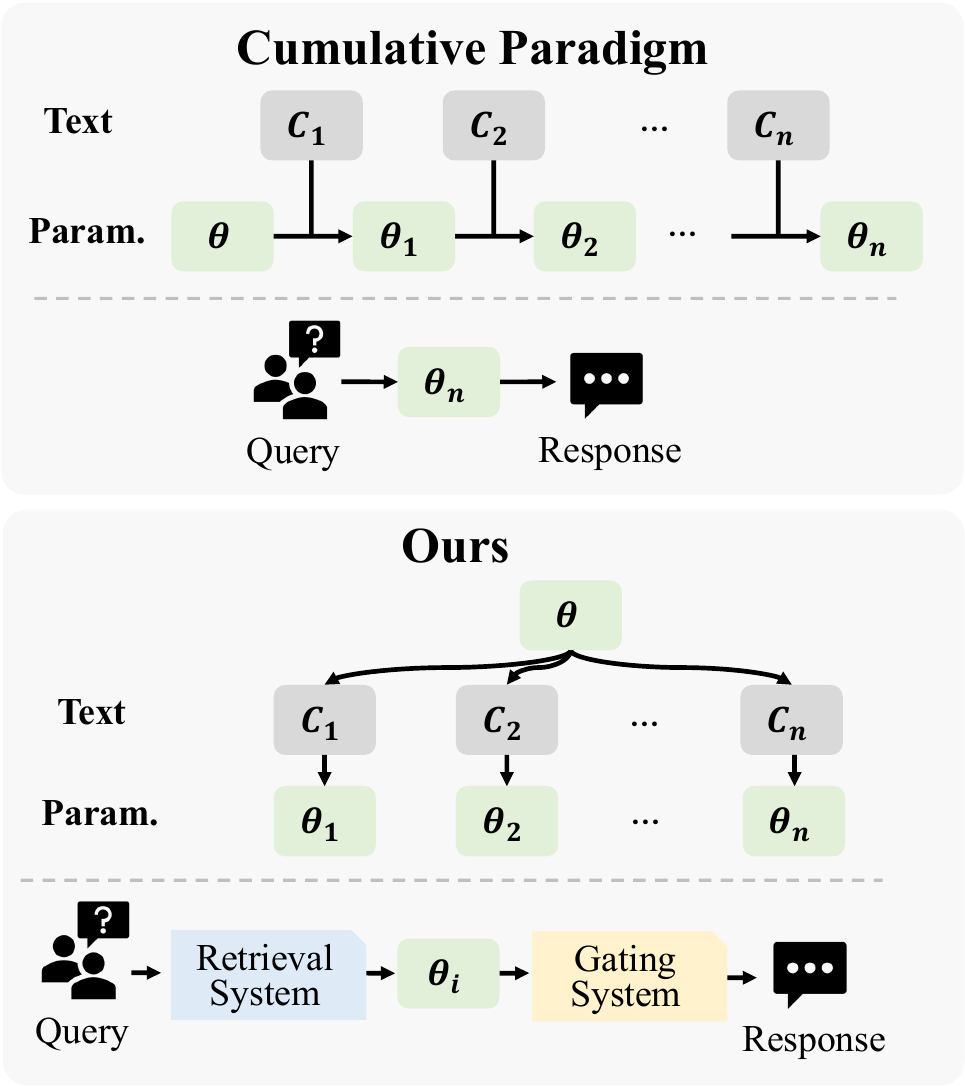}
    \vspace{-6pt}
    \caption{
    Comparison of cumulative context distillation and our method.
    Cumulative distillation compresses all contexts into one adapter, whereas our framework stores contexts as \textit{modular memories}: \textit{retrieves, routes}, and \textit{gates} the latent memories for each query.
    }
    \vspace{-20pt}
    \label{fig:comparsion}
\end{figure}
However, the second part of the question remains largely unaddressed. 
Practical deployment requires not only writing a context into parameters, but also managing a collection of parameterized memories.
Existing CD methods typically bypass this issue by assuming oracle access to the relevant distilled memory~\citep{ye2026OPCD,zhang2026opsdl,snell2022learning,caccia2025training}, or by continually writing new contexts into a single adapter with a cumulative paradigm~\citep{wang2024templora,cao2025infiniteicl}. 

The cumulative paradigm, as illustrated in Figure~\ref{fig:comparsion}, is conceptually simple: each new context is absorbed into the current parameter state, so the resulting model is expected to implicitly contain all previously observed contexts. 
As a result, newly distilled contexts can overwrite earlier ones~\citep{shenfeld2026SDFT,li2017forgetting,kirkpatrick2017forgetting}, leading to severe forgetting and unreliable activation of previously latent memories. 
Moreover, since all contexts are compressed into a single state, the model lacks an explicit mechanism to select among memories or to deactivate them when a query does not require contextual knowledge. 
Consequently, cumulative distillation is fragile in realistic non-oracle settings, where the system must determine both which latent memory to use and whether any memory should be activated. 
% \todo{cite some work about forgetting}

In this paper, we formulate context distillation as a \emph{latent memory management} problem, where distilled contexts are maintained as persistent modular memories. Specifically, we distill each context into a dedicated LoRA adapter to form a memory bank. As shown in Figure~\ref{fig:comparsion}, our framework retrieves candidate memories for a given query, routes it to the most suitable adapter, and employs a Self-Gating mechanism to decide whether latent memory should be activated, avoiding harmful use on context-agnostic queries. To improve efficiency, we further introduce cache-sharing to amortize context encoding during distillation and reduce retrieval and gating overhead at inference time. Experiments under realistic settings show that our method substantially outperforms cumulative-memory baselines, while Self-Gating and cache-sharing improve robustness and efficiency.

% In this paper, we propose that the main bottleneck of context distillation is not merely how to distill a context, but how to manage distilled contexts as persistent latent memories. 
% We therefore formulate context distillation as a \emph{latent memory management} problem. 
% We distill each context independently into a dedicated LoRA adapter, forming a modular memory bank. 
% As illustrated in Figure~\ref{fig:comparsion}, given a query, our framework first performs external retrieval to identify candidate memories, then uses internal routing to select the most appropriate adapter among them. 
% A Self-Gating mechanism further determines whether any latent memory should be activated, preventing harmful memory use on context-agnostic queries. 
% Finally, to make this pipeline practical, we introduce cache sharing to amortize context encoding during distillation and support low-overhead retrieval and activation at inference time.

% Experiments validate the importance of latent memory management for context distillation. 
% Under realistic setting, our method substantially outperforms cumulative-memory baselines, and Self-Gating prevents harmful activation on context-agnostic queries.
% Moreover, our proposed cache sharing reduces retrieval and gating FLOPs.

Our contributions are threefold:
\begin{itemize}
    \vspace{-7pt}
    \item We formulate context distillation as latent memory management, shifting focus from single-context internalization to storage, retrieval, and activation of latent memories.
    \vspace{-8pt}
    \item We propose a modular adapter memory bank with two-stage latent retrieval and Self-Gating.
    \vspace{-20pt}
    \item We introduce cache-sharing distillation, enabling efficient adapter switching while preserving downstream performance.
\end{itemize}
\vspace{-5pt}

\vspace{-5pt}
\section{Background}
\vspace{-5pt}
% Context Distillation aim to distill the contextual information into latent space, either the latent token through KV-cache compression or the model parameter through context internalization. \todo{should use the Unified Terminology}

\subsection{KV-cache Compression}
KV-cache compression aims to compress long contexts into shorter latent tokens, typically achieving a $15\times$-$30\times$ compression ratio. Previous works~\citep{eyuboglu2025cartridges,liu2025c3,li2026lcc} first compress the context $c$ with a compressor $\phi$ and apply context distillation target:
\vspace{-5pt}
\begin{equation}
    D_{\mathrm{KL}}\!\left(
    \pi(\cdot \mid q,c)
    \,\middle\|\,
    \pi(\cdot \mid q,\phi(c))
    \right),
\end{equation}
% \vspace{-5pt}
where $q$ denotes a query.
The selection of $\phi$ varies across different works. Cartridges~\citep{eyuboglu2025cartridges} and Latent Context Compilation~\citep{li2026lcc} compress context via Base Model with LoRA, while C3~\citep{liu2025c3} uses a small Base Model as context compression encoder.
\vspace{-5pt}
\subsection{Context Distillation} 
Context distillation aims to internalize the context into the model parameters by minimizing:
% \begin{equation}
%     D_{KL}(\pi_{\theta}(\cdot|q,c)  \pi_{\theta+\Delta\theta}(\cdot|q))
% \end{equation}
\vspace{-5pt}
\begin{equation}
    D_{\mathrm{KL}}\!\left(
    \pi_{\theta}(\cdot\mid q,c)
    \,\middle\|\,
    \pi_{\theta+\Delta\theta}(\cdot\mid q)
    \right).
\end{equation}
\citet{snell2022learning} first proposed internalizing contextual information into model parameters through distillation.
Similarly, Prompt baking~\citep{bhargava2024promptbaking} proposes baking prompt information into model parameters.
\citet{charakorn2025text2lora,charakorn2026doc2lora} adopt the idea of meta-learning and use a hypernetwork to facilitate the distillation.
Recently, OPCD~\citep{ye2026OPCD} and Opsdl~\citep{zhang2026opsdl} adopt on-policy distillation to further enhance the distillation capability. 
\citet{caccia2025training} first discuss the synergies between CD and RAG. However, they only augment the oracle latent memory with the retrieved passage, instead of managing the latent memory itself. 
Beyond internalizing a single context, how to manage latent memories remains largely unaddressed.
% Cumulative paradigm, like Temp-Lora~\citep{wang2024templora} and InfiniteICL~\citep{cao2025infiniteicl} propose to internalize a stream of context into a single adapter, i.e. $\theta_{i-1} + c_i \rightarrow \theta_i$.

\begin{figure*}
    \centering
    \includegraphics[width=1.0\linewidth]{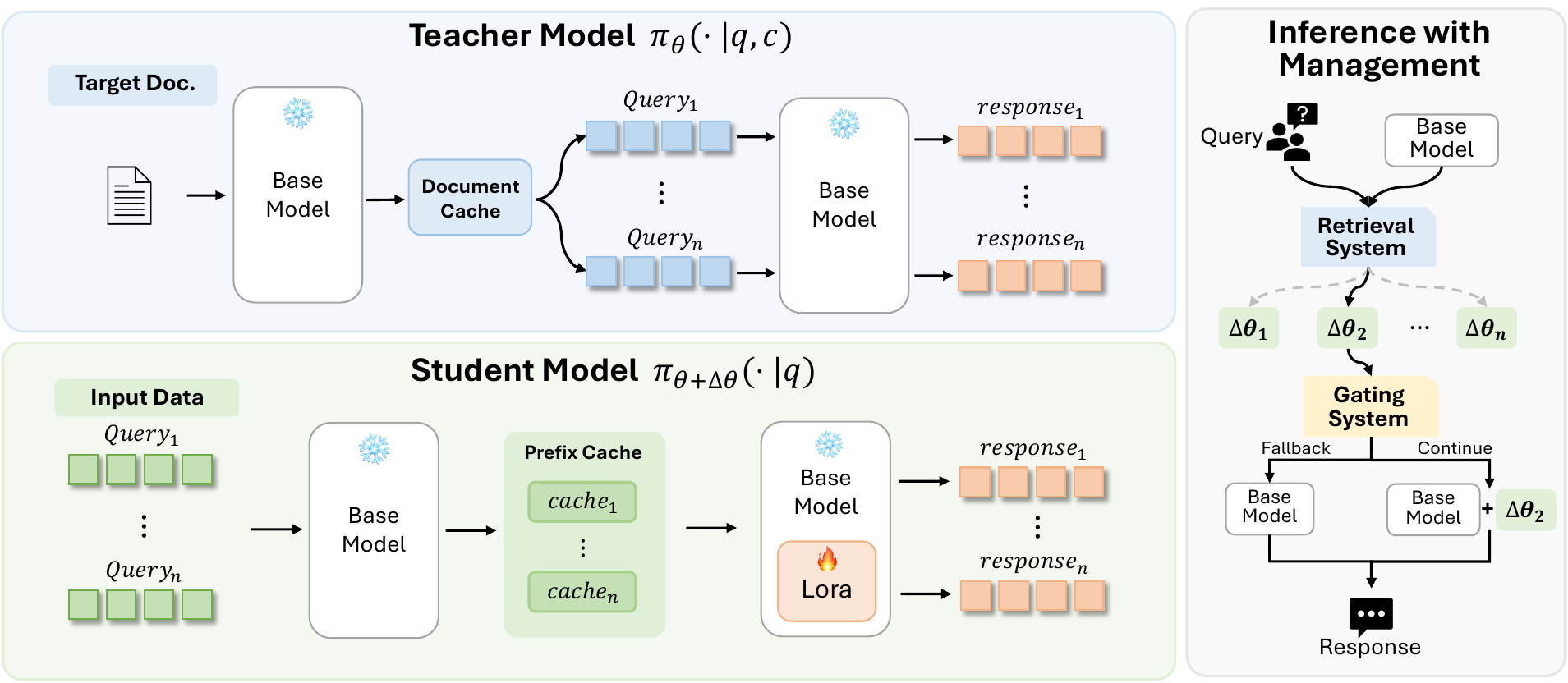}
    \caption{Overview of our training stage and inference stage. \textbf{Training:} Teacher model precomputes the KV-cache of target document for reuse during training, while student computes the prefix KV-cache with Base Model to enable the \textit{Hot-swap} property for management. \textbf{Inference:} Given a query, we first retrieve the most relevant latent memory $\Delta\theta_i$, and then use Self-Gating to decide whether to continue with $\theta+\Delta\theta_i$ or fallback to $\theta$. Owing to the cache-sharing distillation introduced during training, memory management remains efficient. }
    \label{fig:overview}
    \vspace{-10pt}
\end{figure*}

\vspace{-5pt}
\subsection{Cumulative Distillation}
\label{sec:cumulative}
% \todo{cite some forgetting paper}
Cumulative context distillation~\citep{wang2024templora,cao2025infiniteicl} updates the model sequentially by distilling each new context into its parameters:
\begin{equation}
\theta_i = \arg\min_{\theta} D_{\mathrm{KL}}\big(
\pi_{\theta_{i-1}}(\cdot \mid q, c_i) \parallel \pi_{\theta}(\cdot \mid q)
\big),
\end{equation}
with the goal of progressively internalizing all preceding contexts into the latest model $\theta_n$, thereby enabling inference with the latest adapter instead of relying on an oracle selection of the appropriate adapter.
However, such sequential writing can lead to severe catastrophic forgetting~\citep{shenfeld2026SDFT,li2017forgetting,kirkpatrick2017forgetting}. 
Beyond the forgetting issue, we further argue that
under an idealized recursive distillation assumption, cumulative distillation behaves analogously to distilling the Base Model conditioned on accumulated contexts $\{c_1,\ldots,c_i\}$:
\begin{equation}
    \pi_{\theta_i^*}(y \mid q)
    =
    \pi_{\theta}(y \mid q, [c_1,\ldots,c_i]).
\end{equation}
We provide further details in Appendix~\ref{apdx:cumulative}. 
This perspective suggests that cumulative CD inherits the teacher model's sensitivity to irrelevant accumulated contexts, as well as its degradation under long-context inputs. Consequently, performance may deteriorate even under the oracle test setting.
% This indicates that the theoretical upper bound of cumulative distillation is equivalent to directly conditioning the Base Model on the full concatenated context. Therefore, this paradigm cannot fundamentally surpass the Base Model's ability to utilize long contexts, and it still inherits common long-context degradation issues.

\section{Methodology}

\subsection{Cache-Sharing Context Distillation}

% To avoid catastrophic forgetting from cumulative updates, we decouple context assimilation across context chunks. 
Given a context stream $\mathcal{C}=\{c_1,c_2,\dots,c_t\}$, we freeze the base model $f_\theta$ and train a separate PEFT module, e.g., LoRA, $\Delta\theta_i$ for each context $c_i$. The student model $\theta+\Delta\theta_i$ is trained to mimic the teacher model, i.e., the base model explicitly conditioned on $c_i$. For a set of queries $\{q_1,\dots,q_n\}$, we minimize the KL divergence:
% \begin{equation}
% \begin{split}
%     \Delta\theta_i^* = \mathop{\arg\min}_{\Delta\theta_i} \frac{1}{n}\sum_{j=1}^n D_{\text{KL}} \big( & \pi_{\theta}(y_j|q_j, c_i) \\
%     & \parallel \pi_{\theta+\Delta\theta_i}(y_j|q_j) \big),
% \end{split}
% \label{eq:distillation}
% \end{equation}
\begin{equation}
\begin{split}
    \Delta\theta_i^* = \mathop{\arg\min}_{\Delta\theta_i} \frac{1}{n}\sum_{j=1}^n D_{\text{KL}} \big( & \pi_{\theta}(y_j|q_j, c_i) \\
    & \parallel \pi_{\theta+\Delta\theta_i}(y_j|\text{KV}_{q_j}) \big),
\end{split}
\label{eq:distillation}
\end{equation}
where $\text{KV}_{q_j}=f_\theta(q_j)$ and $y_j$ denotes the target response distribution produced by the teacher.
To reduce both training and inference cost, we introduce a dual cache-sharing mechanism for the teacher and student models.
\vspace{-5pt}
\paragraph{Teacher Cache}  
For the teacher $\pi_{\theta}(\cdot|q_j,c_i)$, prefilling the long context $c_i$ dominates computation. During distillation of a fixed adapter $\Delta\theta_i$, however, $c_i$ remains unchanged, while only $q_j$ and $y_j$ vary. We therefore precompute KV-cache of $c_i$ once with the base model and reuse it for all training steps, substantially accelerating teacher-side generation.
\vspace{-5pt}
\paragraph{Student Cache} 
We further introduce a decoupled cache-sharing pipeline for the student, yielding nearly zero-overhead adapter switching, which we refer to as  the \textbf{\textit{Hot-swap}} property. 
Specifically, student-side cache sharing deliberately trains a cache-compatible student distribution. Instead of applying the adapter while encoding the query prefix, we compute the prefix KV-cache with the frozen base model and activate the adapter only for response generation. This restricts the adapter to be compatible with base-model prefix caches, enabling adapter hot-swapping without KV recomputation. 
Consequently, the system can seamlessly switch among adapters $\Delta\theta_i$ while reusing the same prefix cache, which is essential for the Gating and Retrieval systems described in Sections~\ref{sec:gating} and~\ref{sec:retrieval}.

\subsection{Self-Gating System}
\label{sec:gating}

\begin{figure}[t]
    \centering
    \includegraphics[width=1.0\linewidth]{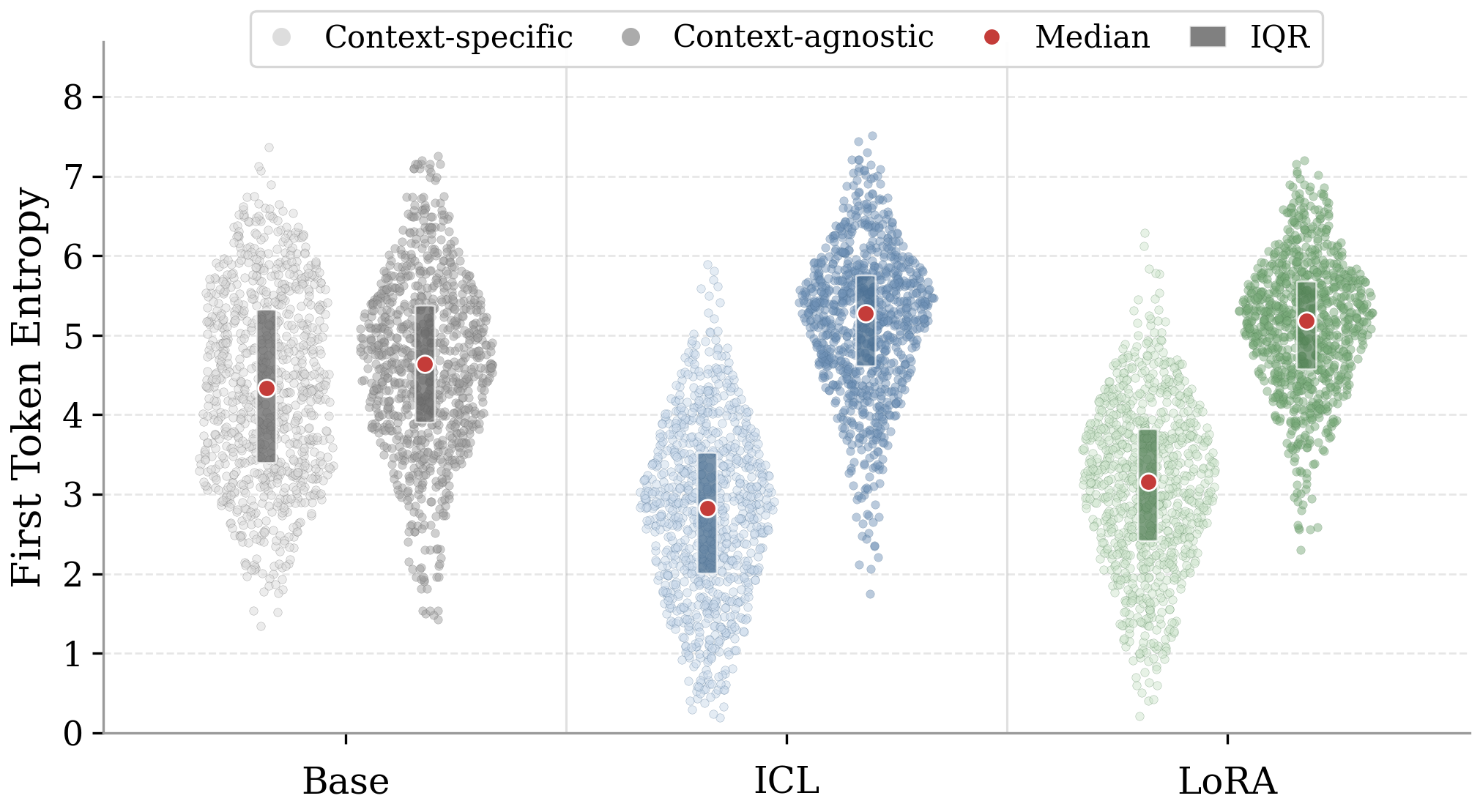}
    \vspace{-15pt}
    \caption{Distribution of the first-token entropy. A clear gap exists between context-specific and context-agnostic queries when context is provided, whereas the base model exhibits no such distinction.}
    \label{fig:entropy_distribution}
    \vspace{-10pt}
\end{figure}

As illustrated in Figure~\ref{fig:entropy_distribution}, we observe a distinct phenomenon in the predictive entropy of the first generated token. In standard In-Context Learning (ICL), there exists a clear distributional margin between context-specific queries and context-agnostic queries. In contrast, the base model without any appended context exhibits no such distributional gap. Since our distillation objective in Eq.~\ref{eq:distillation} explicitly aligns the output distribution of the LoRA student with the ICL teacher, the trained context-specific LoRA models inherit this entropy gap.

\begin{figure}[t]
    \centering
    \includegraphics[width=1.0\linewidth]{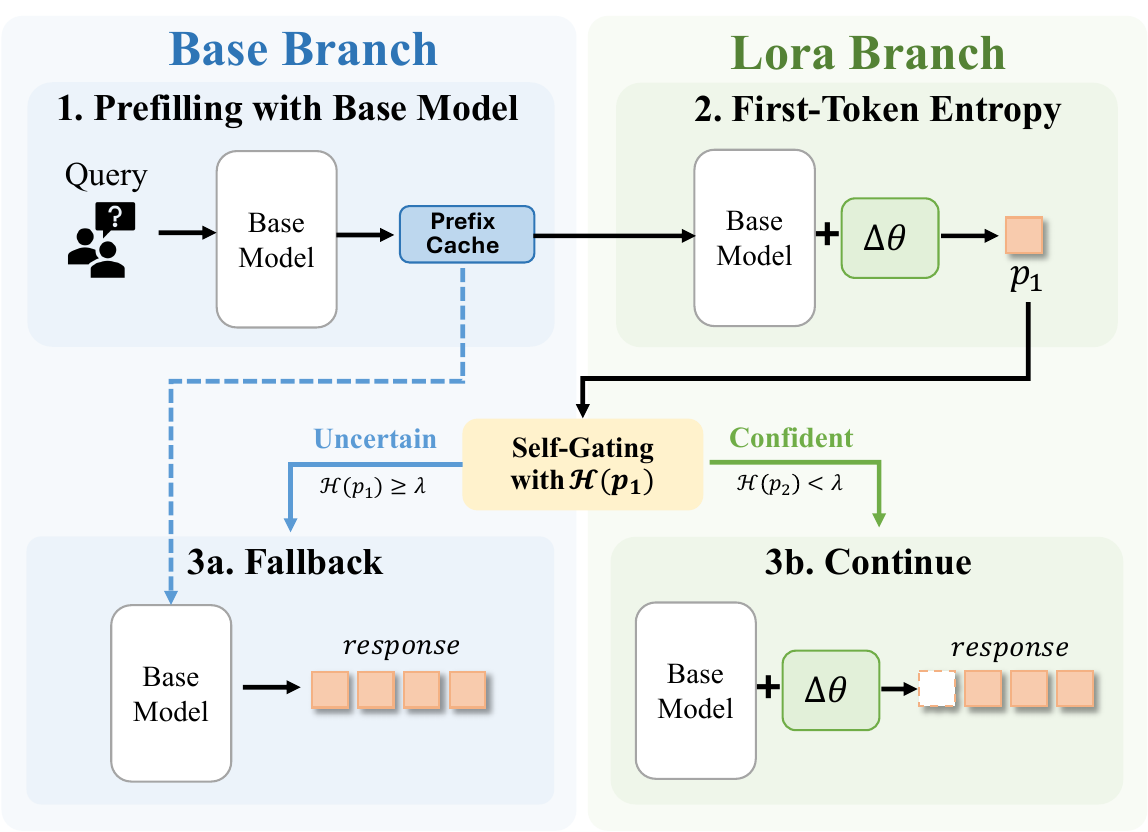}
    \caption{Overview of the proposed Self-Gating mechanism. The system routes the generation between the specific LoRA and the base model based on the first-token entropy, with zero KV-cache recomputation.}
    \label{fig:gating}
    \vspace{-5pt}
\end{figure}

Computing this first-token entropy with standard ICL is prohibitively expensive due to long-context. However, our cache-sharing mechanism makes it computationally trivial. Motivated by this observation, we propose a \textit{Self-Gating} mechanism that leverages the first-token entropy for dynamic model gating. As shown in Figure~\ref{fig:gating}, given a context adapter $\Delta\theta_i$, we first process the input prefix (i.e., prompt and query, denoted simply as $q$), exclusively through the frozen base model $f_{\theta}$ to obtain the shared prefix KV-cache:
\vspace{-5pt}
\begin{equation}
    \text{KV}_{q} = f_{\theta}(q).
\end{equation}
We then temporarily activate the LoRA module $\Delta\theta_i$ to generate the probability distribution $p_1$ of the first token:
\vspace{-5pt}
\begin{equation}
    p_1 = f_{\theta+\Delta\theta_i}(\text{KV}_{q}).
\label{eq:first_token}
\end{equation}

Let $\mathcal{H}(p_1)$ denote the Shannon entropy of this distribution. We use a threshold $\lambda$ for gating:
\begin{itemize}
    \vspace{-3pt}
    \item \textbf{Continue ($\mathcal{H}(p_1) < \lambda$):} Low entropy indicates that the LoRA model is highly confident in answering the query based on its injected knowledge. We therefore retain the LoRA module and continue decoding.
    \vspace{-5pt}
    \item \textbf{Fallback ($\mathcal{H}(p_1) \ge \lambda$):} High entropy indicates significant uncertainty, suggesting that the query is likely context-agnostic, e.g., a general knowledge question. We then instantly deactivate the LoRA module and fall back to the base model.
    \vspace{-3pt}
\end{itemize}

When falling back, the base model seamlessly reuses the already computed $\text{KV}_{q}$ for subsequent generation. Therefore, the \textit{only} additional computational overhead introduced by Self-Gating is the forward pass of a single token using the LoRA model in Eq.~\ref{eq:first_token}, making the entire gating process exceptionally efficient.

\subsection{Latent Memory Retrieval System}
\label{sec:retrieval}

\begin{figure}[t]
    \centering
    \includegraphics[width=1.0\linewidth]{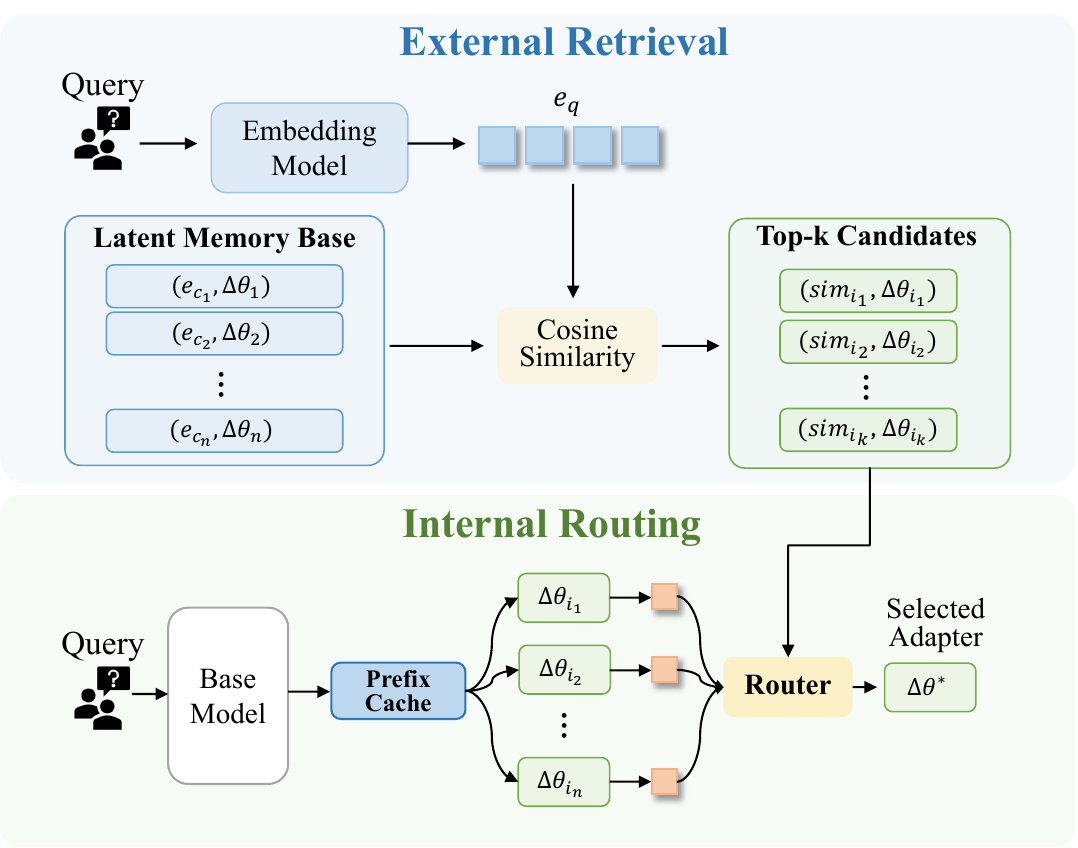}
    \caption{
Latent memory retrieval with independently distilled LoRA adapters.
An external retriever first selects candidate adapters by cosine similarity, and an internal router further ranks them using first-token entropy and hidden states under a shared prefix KV-cache.
}
    \label{fig:rag}
    \vspace{-10pt}
\end{figure}

By distilling each context into an independent LoRA module $\{\Delta\theta_1,\dots,\Delta\theta_t\}$, we recast context utilization as a module selection and routing problem. To manage these distributed latent memories, we introduce a two-stage retrieval system: \textit{External Retrieval} and \textit{Internal Routing}, as shown in Figure~\ref{fig:rag}.

\paragraph{External Retrieval (Coarse-Grained)}
Similar to dense retrieval in RAG, we use an off-the-shelf embedding model $f_{\text{emb}}$ to encode each context and the incoming query:
\vspace{-5pt}
\begin{align}
    \mathbf{e}_{c_i} &= f_{\text{emb}}(c_i), \\
    \mathbf{e}_{q} &= f_{\text{emb}}(q).
\end{align}
We then select the top-$K$ LoRA modules with the highest cosine similarity to the query:
\vspace{-5pt}
\begin{equation}
    \mathcal{S}_{\text{topK}} =
    \mathop{\text{TopK}}_{i \in \{1,\dots,t\}}
    \big( \cos(\mathbf{e}_{c_i}, \mathbf{e}_{q}) \big).
\end{equation}

\paragraph{Internal Routing (Fine-Grained)}
Although external embeddings capture semantic relevance, they may not align perfectly with the LLM generative confidence. We therefore refine the candidate set $\mathcal{S}_{\text{topK}}$ with an internal router.
Using the cache-sharing mechanism in Section~\ref{sec:gating}, we first compute the shared prefix KV-cache with the base model. For each candidate adapter $\Delta\theta_i \in \mathcal{S}_{\text{topK}}$, we briefly activate it to obtain the first-token hidden state $h_i$ and predictive entropy $\mathcal{H}_i$. The final adapter is selected by a lightweight feed-forward network(FFN) router conditioned on the external similarity $sim_i=\cos(\mathbf{e}_{q},\mathbf{e}_{c_i})$, entropy $\mathcal{H}_i$, and hidden state $h_i$ (details in Appendix~\ref{apdx:router-training}):
\begin{equation}
    \Delta\theta^* =
    \arg\max_{\Delta\theta_i \in \mathcal{S}_{\text{topK}}}
    router(sim_i, \mathcal{H}_i, h_i).
\end{equation}
Since all candidates share the same query-prefix KV-cache, evaluating multiple LoRAs during internal routing introduces negligible overhead.

\newcommand{\variant}[1]{\hspace{1.0em}-- #1}
\newcommand{\best}[1]{\textbf{#1}}
\newcommand{\second}[1]{\underline{#1}}

\section{Experiments}

In this section, we empirically study the central question: 
\textit{How should latent memories be stored, retrieved, and safely activated?}
% We organize our experiments around three aspects of latent-memory systems. 
% First, in Section~\ref{sec:mem_store}, we examine the limitations of storing memories through a cumulative paradigm. Second, in Section~\ref{sec:exp_retrieve}, we evaluate whether explicit adapter retrieval enables practical use of latent memories. Finally, in Section~\ref{sec:activate}, we study how latent memories can be safely activated in the presence of both context-specific and context-agnostic queries.

\subsection{Implementation Details}
% \paragraph{Hyperparameters.}
% We use LoRA for context distillation. Unless otherwise specified, we set the LoRA rank to 64 and the LoRA scaling factor $\alpha$ to 128. Each adapter is trained for 10 epochs with a learning rate of $5 \times 10^{-4}$.
\paragraph{Base Model}
We conduct experiments with Qwen2.5-0.5B~\citep{yang2024qwen2.5} and Qwen2.5-7B~\citep{yang2024qwen2.5} as base models. We use Qwen3-Embedding-0.6B~\citep{zhang2025qwen3emb} for retrieval.
\vspace{-5pt}
\paragraph{Benchmarks}
For context-specific queries, we evaluate on NarrativeQA~\citep{kovcisky2018narrativeqa} and SQuAD~\citep{rajpurkar2016squad}. For context-agnostic queries, we use CommonsenseQA~\citep{talmor2019commonsenseqa}. Since SQuAD contains a large number of documents, we retain the top 300 most frequently queried documents, ranked by the number of associated queries. For the statistics of our dataset, please refer to Appendix~\ref{apdx:dataset}.
\vspace{-5pt}
\paragraph{Baselines} We compare with two cumulative-memory baselines, TempLora~\citep{wang2024templora} and InfiniteICL~\citep{cao2025infiniteicl}. 
Since TempLora does not use QA distillation, we also include TempLoraCD, which replaces its original objective with distillation loss for a fairer comparison. 

% To further demonstrate the generalization of our method, we integrate with various distillation methods in previous works~\citep{shenfeld2026SDFT,hubotter2026SDPO,ye2026OPCD} in Section~\ref{sec:generalization} and Appendix~\ref{apdx:exp_distill}.

More details can be found in Appendix~\ref{adpx:exp_detail}.

\begin{figure}[t]
    \centering

    \begin{subfigure}{1.0\linewidth}
        \centering
        \includegraphics[width=1.0\linewidth]{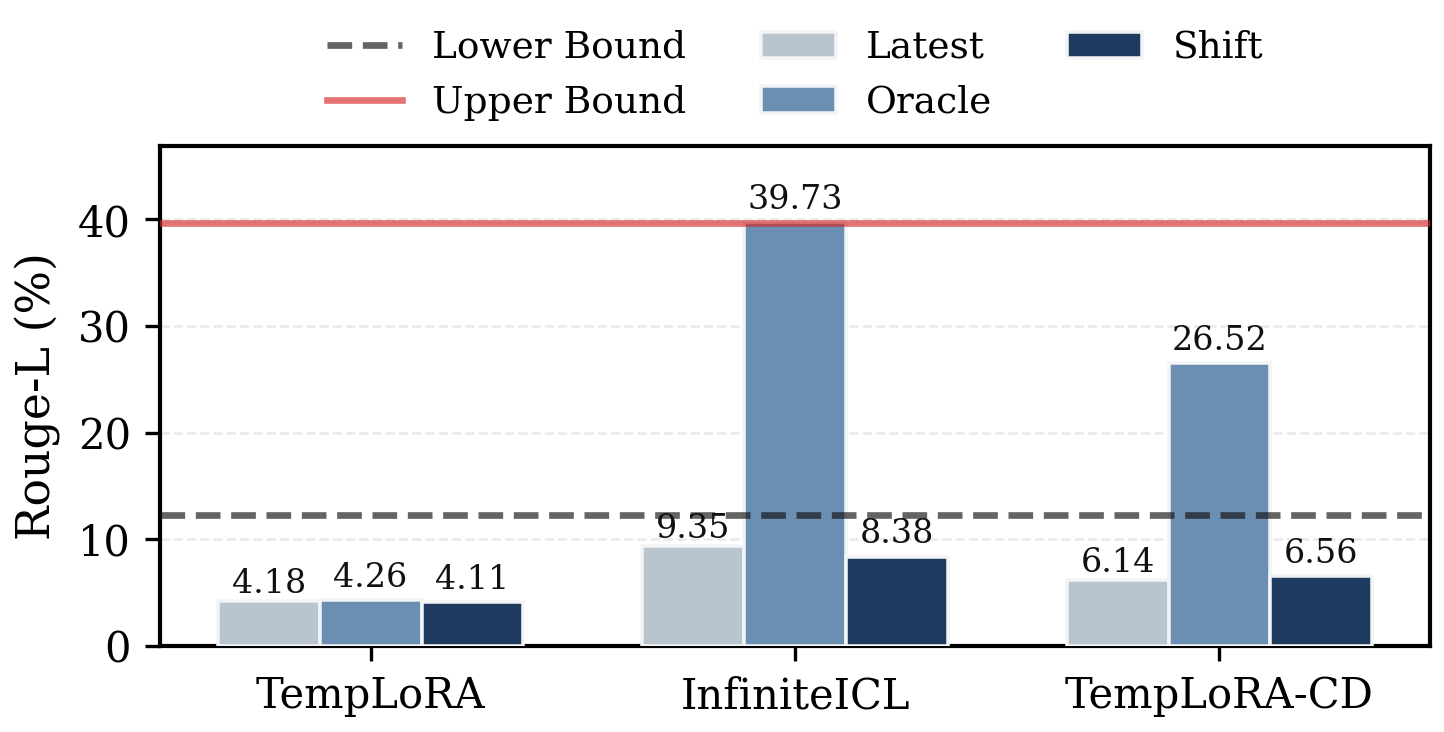}
        \vspace{-20pt}
        \caption{Results on NarrativeQA}
        \label{fig:cumulative_nqa}
    \end{subfigure}
    \begin{subfigure}{1.0\linewidth}
        \centering
        \includegraphics[width=1.0\linewidth]{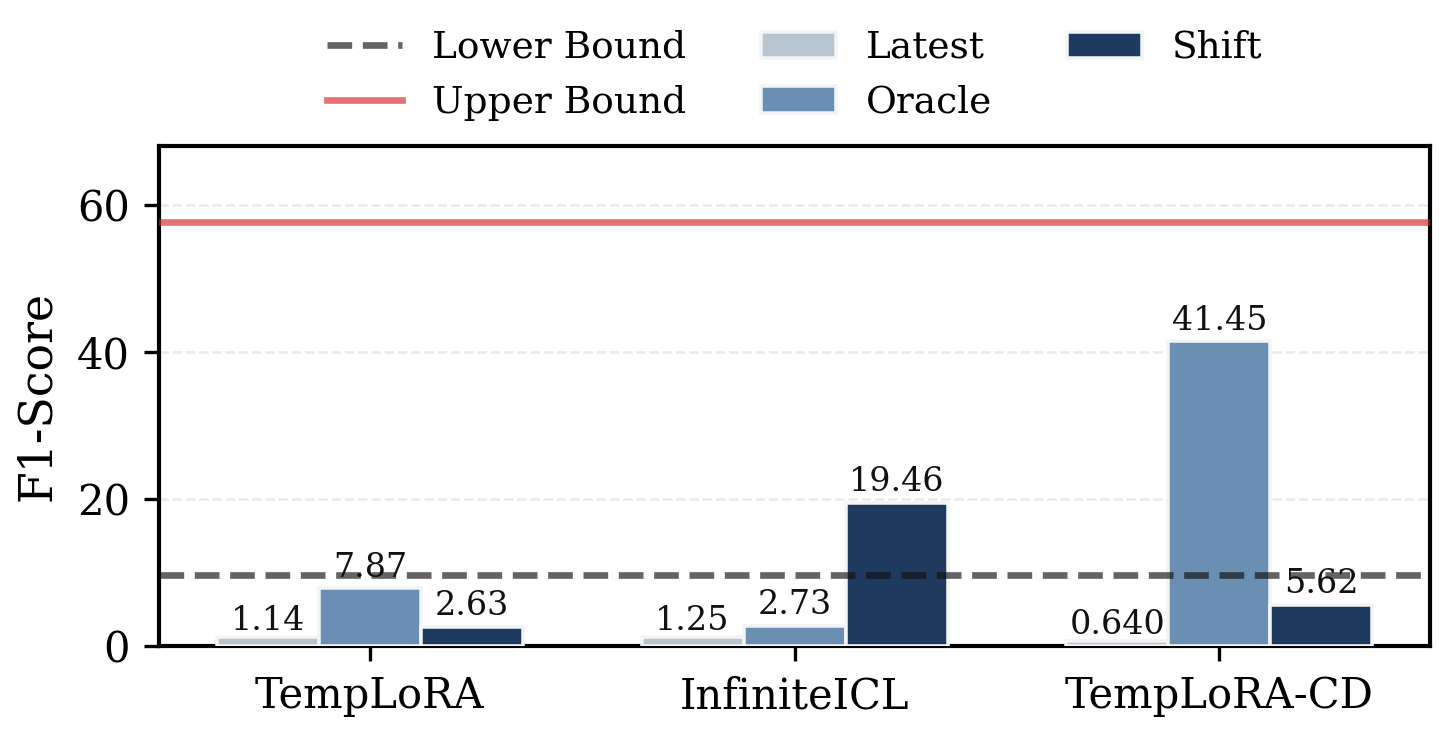}
        \vspace{-20pt}
        \caption{Results on SQuAD}
        \label{fig:cumulative_squad}
    \end{subfigure}
    \vspace{-20pt}
    \caption{Results of cumulative methods on Qwen2.5-0.5B, under various evaluation settings.}
    \label{fig:cumulative_nqa_squad}
    \vspace{-15pt}
\end{figure}

\begin{table*}[t]
\centering
\small
\setlength{\tabcolsep}{5.2pt}
\renewcommand{\arraystretch}{1.14}
\begin{threeparttable}
\caption{
Main results on NarrativeQA and SQuAD.
 $\dagger$ denotes Oracle settings with access to the ground-truth context or adapter, serving as upper bounds for the corresponding methods.
}
\vspace{-5pt}
\label{tab:main_results_qwen}

\begin{tabular}{@{}lcccccccc@{}}
\toprule
\multirow{3}{*}{Method}
& \multicolumn{4}{c}{NarrativeQA}
& \multicolumn{4}{c}{SQuAD} \\
\cmidrule(lr){2-5} \cmidrule(lr){6-9}
& \multicolumn{2}{c}{\textsc{Qwen2.5-0.5B}}
& \multicolumn{2}{c}{\textsc{Qwen2.5-7B}}
& \multicolumn{2}{c}{\textsc{Qwen2.5-0.5B}}
& \multicolumn{2}{c}{\textsc{Qwen2.5-7B}} \\
\cmidrule(lr){2-3} \cmidrule(lr){4-5} \cmidrule(lr){6-7} \cmidrule(lr){8-9}
& ROUGE-1 $\uparrow$
& ROUGE-L $\uparrow$
& ROUGE-1 $\uparrow$
& ROUGE-L $\uparrow$
& EM $\uparrow$
& F1 $\uparrow$
& EM $\uparrow$
& F1 $\uparrow$ \\
\midrule

\rowcolor{gray!10}
\multicolumn{9}{@{}l}{\textbf{Lower Bound}} \\
\addlinespace[2pt]
Base Model
& 13.08 & 12.26 & 14.95 & 14.40
& 2.38 & 9.63 & 13.51 & 21.97 \\

\addlinespace[3pt]
\rowcolor{gray!10}
\multicolumn{9}{@{}l}{\textbf{Upper Bound}} \\
\addlinespace[2pt]
\graytext{In-context$^{\dagger}$}
& \graytext{40.65} & \graytext{39.64} & \graytext{55.99} & \graytext{55.03}
& \graytext{41.98} & \graytext{57.67} & \graytext{78.20} & \graytext{85.40} \\

RAG
& & & & & & & & \\

\variant{@3}
& 21.12 & 20.21 & 47.36 & 46.61
& 26.84 & 42.61 & 63.89 & 72.18 \\

\variant{@5}
& 16.66 & 15.79 & 35.57 & 34.87
& 26.04 & 41.15 & 65.18 & 73.37 \\

\addlinespace[3pt]
\rowcolor{gray!10}
\multicolumn{9}{@{}l}{\textbf{Cumulative Paradigm}} \\
\addlinespace[2pt]
TempLora
& 4.67 & 4.18 & 4.83 & 4.11
& 0.00 & 1.14 & 0.00 & 1.60 \\

\addlinespace[2pt]
InfiniteICL
& 10.10 & 9.35 & 4.24 & 3.77
& 0.76 & 1.25 & 0.91 & 0.91 \\

\addlinespace[2pt]
TempLoraCD
& 6.65 & 6.14 & 4.09 & 3.78
& 0.00 & 0.64 & 0.00 & 0.00 \\

\addlinespace[3pt]
\rowcolor{gray!10}
\multicolumn{9}{@{}l}{\textbf{Ours}} \\
\addlinespace[2pt]
\graytext{Oracle$^{\dagger}$}
& \graytext{31.23} & \graytext{30.22} & \graytext{37.55} & \graytext{36.78}
& \graytext{33.38} & \graytext{50.05} & \graytext{43.32} & \graytext{56.97} \\

\addlinespace[2pt]
Retrieval
& & & & & & & & \\

\variant{@3}
& 24.54 & 23.57 & 28.46 & 27.68
& \textbf{28.30} & 43.11 & \textbf{36.34} & \textbf{46.57} \\

\variant{@5}
& \textbf{24.44} & \textbf{23.46} & \textbf{28.64} & \textbf{27.86}
& 28.14 & \textbf{43.25} & 36.30 & 46.45 \\

\bottomrule
\end{tabular}

\end{threeparttable}
\vspace{-10pt}
\end{table*}

\subsection{How to Store}

\label{sec:mem_store}

In this section, we evaluate cumulative paradigm with different settings. The \textit{Latest} setting always uses the latest adapter $\Delta \theta_n$, the \textit{Shift} setting evaluates a query $q^i$ from document $c_i$ with the next adapter $\Delta \theta_{i+1}$, and \textit{Oracle} settings assume access to the correct context or adapter. More details can be found in Appendix~\ref{apdx:baseline_eval}. 

We discuss three key limitations of the cumulative paradigm.
First, the cumulative paradigm suffers from severe forgetting in realistic settings.
Cumulative methods degrade substantially in both \textit{Latest} and \textit{Shift} settings.
For example, TempLoraCD drops to 6.56 ROUGE-L on NarrativeQA and 5.62 EM on SQuAD under \textit{Shift} setting, despite much stronger performance in the \textit{Oracle} setting.
This indicates that despite the cumulative distillation, the latent memory is still highly sensitive to adapter selection and prone to retrieval errors.

Second, the cumulative paradigm limits the upper bound of context distillation.
Even under $Oracle$ setting, cumulative training is affected by increasing cumulative context length and accumulated noise, as discussed in Section~\ref{sec:cumulative}.
As a result, TempLoraCD underperforms our method in Table~\ref{tab:main_results_qwen}: on NarrativeQA, it achieves 26.52 ROUGE-L compared to 30.22 for ours; on SQuAD, it achieves 41.45 F1-Score compared to 50.05 for ours.
This suggests that accumulating contexts weakens an adapter's ability to preserve context-specific knowledge, as discussed in Section~\ref{sec:cumulative}.

Last, the cumulative paradigm fails to manage the memory usage.
InfiniteICL shows unstable behavior: it is unclear whether predictions come from latent adapter memory or explicit context. 
For a query $q^i$ from document $c_i$, InfiniteICL uses $\Delta \theta_{i-1}$ as long-term memory and $c_i$ as short-term memory.
On NarrativeQA, it relys more on short-term context, performing well in the \textit{Oracle} setting but dropping from 39.73 to 8.38 ROUGE-L under \textit{Shift}.
On SQuAD, the trend is reversed: F1-Score increases from 2.73 to 19.46 under \textit{Shift}, indicating greater reliance on latent memory in adapter.
Such results demonstrate that InfiniteICL lacks a stable mechanism to control when to use latent memory versus explicit context.
These results motivate explicit adapter management, rather than relying on a single cumulatively updated memory.

\begin{table*}[t]
\centering
\small
\setlength{\tabcolsep}{6pt}
\renewcommand{\arraystretch}{1.12}
\begin{threeparttable}
\caption{
Results of hybrid queries from NarrativeQA and CommonsenseQA.
Best overall results are shown in \textbf{bold}, and second-best results are \underline{underlined}.
}

\label{tab:gating_results}
\vspace{-20pt}
\begin{tabular}{@{}lcccccc@{}}
\toprule
\multirow{3}{*}{Setting}
& \multicolumn{4}{c}{NarrativeQA}
& \multicolumn{2}{c}{CommonsenseQA} \\
\cmidrule(lr){2-5} \cmidrule(lr){6-7}
& \multicolumn{2}{c}{\textsc{Qwen2.5-0.5B}}
& \multicolumn{2}{c}{\textsc{Qwen2.5-7B}}
& \textsc{Qwen2.5-0.5B}
& \textsc{Qwen2.5-7B} \\
\cmidrule(lr){2-3} \cmidrule(lr){4-5} \cmidrule(lr){6-6} \cmidrule(lr){7-7}
& ROUGE-1 $\uparrow$
& ROUGE-L $\uparrow$
& ROUGE-1 $\uparrow$
& ROUGE-L $\uparrow$
& Accuracy $\uparrow$
& Accuracy $\uparrow$ \\
\midrule

% \multicolumn{7}{@{}l}{\textit{Baselines}} \\
Base Model        
& 13.08
& 12.26
& 14.95
& 14.40
& \textbf{47.80\%}
& \textbf{85.18\%} \\

TempLora          
& 4.67
& 4.18
& 4.83
& 4.11
& 20.56\%
& 20.31\% \\

InfiniteICL       
& 10.10
& 9.35
& 3.93
& 3.56
& 19.49\%
& 20.56\% \\

TempLoraCD        
& 6.65
& 6.14
& 0.75
& 0.67
& 19.57\%
& 18.92\% \\
\midrule

Ours w/o Gating       
& \textbf{23.47}
& \textbf{22.54}
& \underline{26.36}
& \underline{25.67}
& 37.30\%
& 72.40\% \\

Ours
& \underline{22.74}
& \underline{21.84}
& \textbf{26.98}
& \textbf{26.30}
& \underline{45.62\%}
& \underline{79.44\%} \\
\bottomrule
\end{tabular}
\end{threeparttable}
\vspace{-10pt}
\end{table*}

\subsection{Which to Use}
\label{sec:exp_retrieve}

The results in Section~\ref{sec:mem_store} show that cumulative memories are difficult to use when the correct adapter is unknown. 
We therefore evaluate whether our proposed retrieval system can make latent memories practical at inference time. 

As shown in Table~\ref{tab:main_results_qwen}, retrieved adapters consistently outperform cumulative baselines under realistic settings. 
On NarrativeQA, our method achieves 24.54 ROUGE-1 with \textsc{Qwen2.5-0.5B} and 28.64 ROUGE-1 with \textsc{Qwen2.5-7B}, substantially higher than the \textit{latest} variants of TempLora, InfiniteICL, and TempLoraCD. 
On SQuAD, our Retrieval@3 setting reaches 28.30 EM / 43.11 F1 with \textsc{Qwen2.5-0.5B}, and 36.34 EM / 46.57 F1 with \textsc{Qwen2.5-7B}. 
These results indicate that independently stored adapters can function as useful latent memories when paired with a retrieval system.

We further compare our method with the oracle setting. Although the retrieved adapters do not always match oracle performance, they provide a deployable alternative when the query--adapter correspondence is unknown. Moreover, compared with the cumulative method, whose performance drops significantly when moving from the oracle to the realistic setting, our retrieval system preserves a substantial fraction of oracle performance while removing the need for access to the oracle adapter.

\begin{figure}[t]
    \centering

    \begin{subfigure}{0.9\linewidth}
        \centering
        \includegraphics[width=0.9\linewidth]{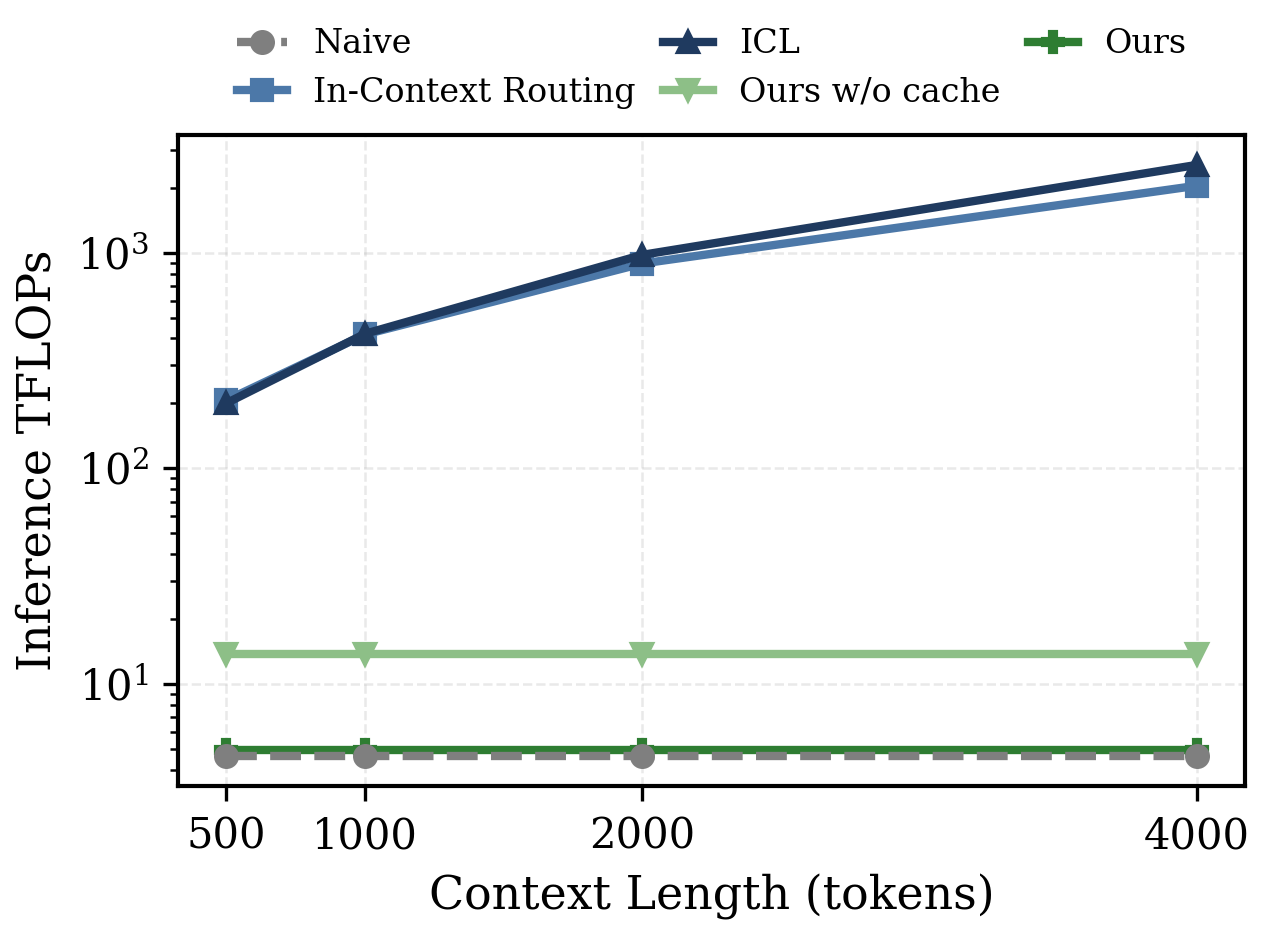}
        \vspace{-6pt}
        \caption{Top-3}
        \label{fig:rag3}
    \end{subfigure}

    \begin{subfigure}{0.9\linewidth}
        \centering
        \includegraphics[width=0.9\linewidth]{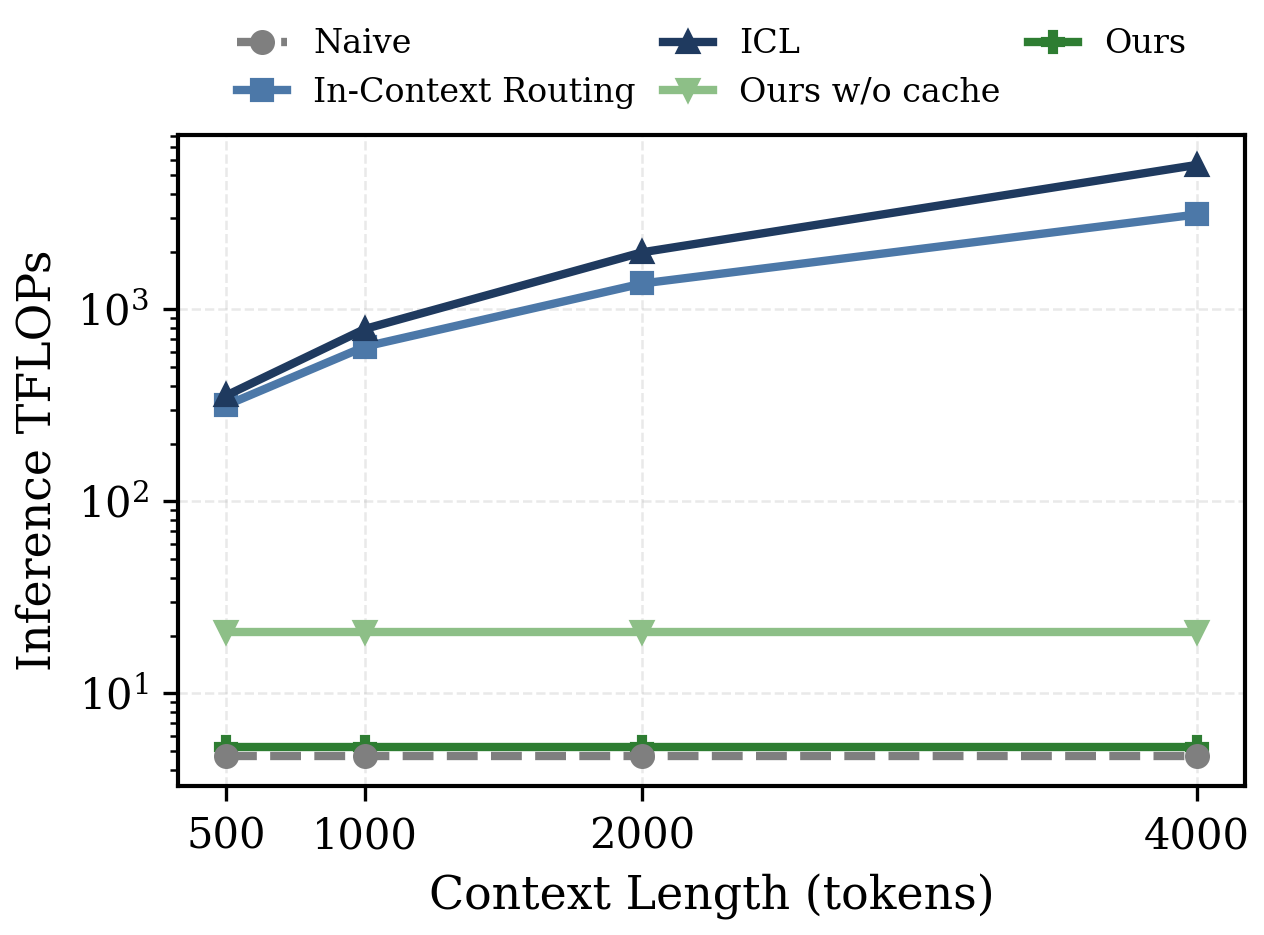}
        \vspace{-6pt}
        \caption{Top-5}
        \label{fig:rag5}
    \end{subfigure}
    \vspace{-10pt}
    \caption{Inference efficiency comparison for retrieval.}
    \label{fig:rag_3_5}
    \vspace{-20pt}
\end{figure}

\vspace{-8pt}
\paragraph{Efficiency}
Compared with standard textual RAG, our goal is not to universally outperform explicit retrieval in task performance, but to build a retrieval system that manages latent memories and makes them effective in realistic settings at low cost.
Figure~\ref{fig:rag_3_5} compares inference efficiency under different retrieval settings. 
\textit{ICL} denotes generation with concatenated retrieved contexts, while \textit{naive} denotes LoRA-based generation without management. \textit{In-Context Routing} routes using the first token generated with context, and \textit{Ours w/o cache} routes with LoRA but without the shared prefix cache. 
More details are provided in Appendix~\ref{apdx:efficiency}. 
Explicit in-context methods process retrieved contexts as part of the input, causing FLOPs to grow rapidly with context length. In contrast, our method switches LoRA adapters with the shared prefix cache, avoiding repeated computation over retrieved contexts and keeping the cost close to naive generation. 
At 4000 tokens, our method reduces FLOPs by $414.7\times$ under Top-3 and $1081.4\times$ under Top-5 compared with explicit in-context inference. Cache reuse further brings a $2.8\times$ and $4.0\times$ reduction over the w/o-cache variant under Top-3 and Top-5, respectively.

\vspace{-5pt}

\begin{figure*}[h]
    \centering
    \begin{subfigure}[t]{0.32\linewidth}
        \centering
        \includegraphics[width=\linewidth]{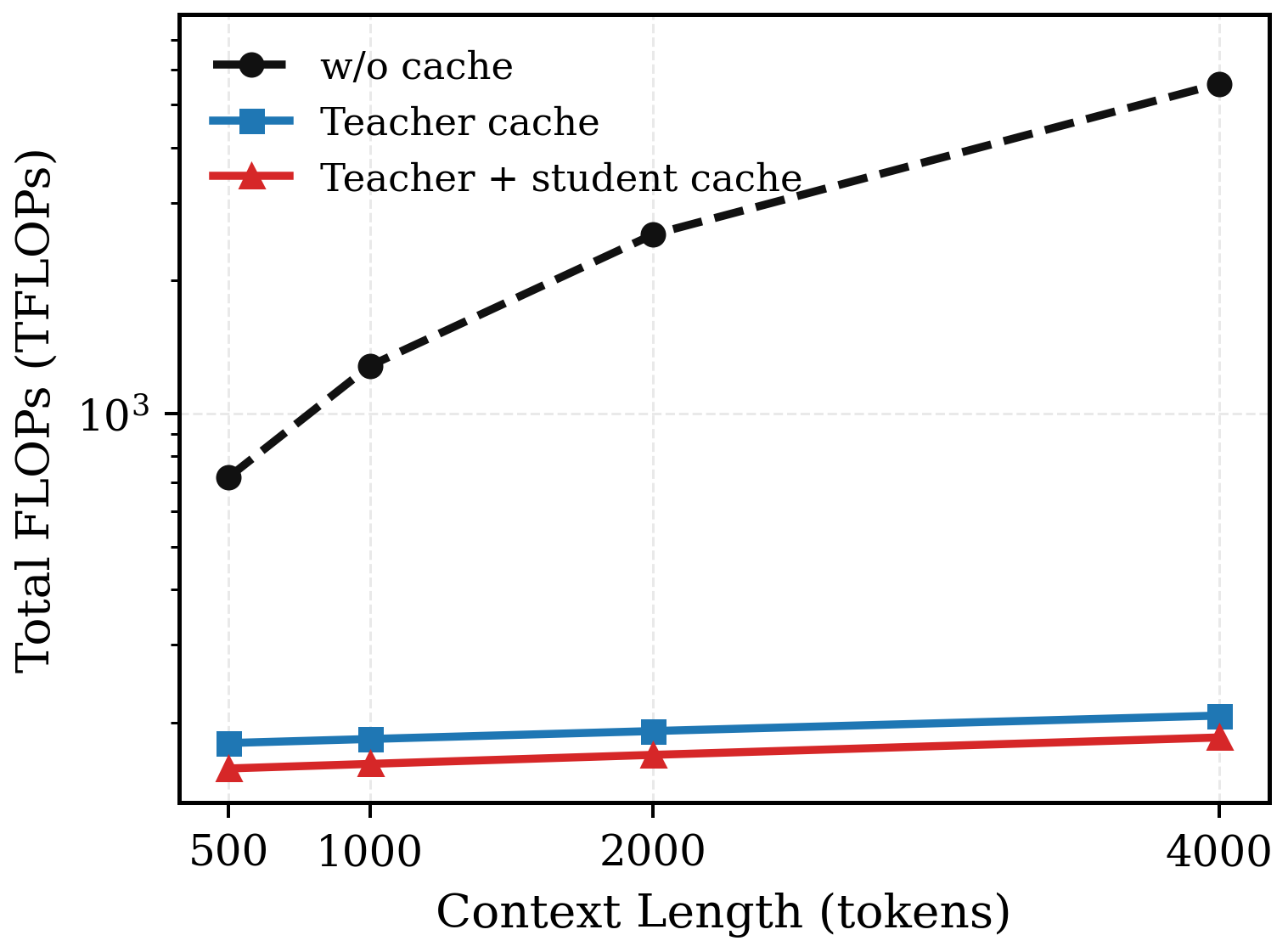}
        \vspace{-20pt}
        \caption{Flops Comparison}
        % \label{fig:fig1}
    \end{subfigure}
    \hfill
    \begin{subfigure}[t]{0.32\linewidth}
        \centering
        \includegraphics[width=\linewidth]{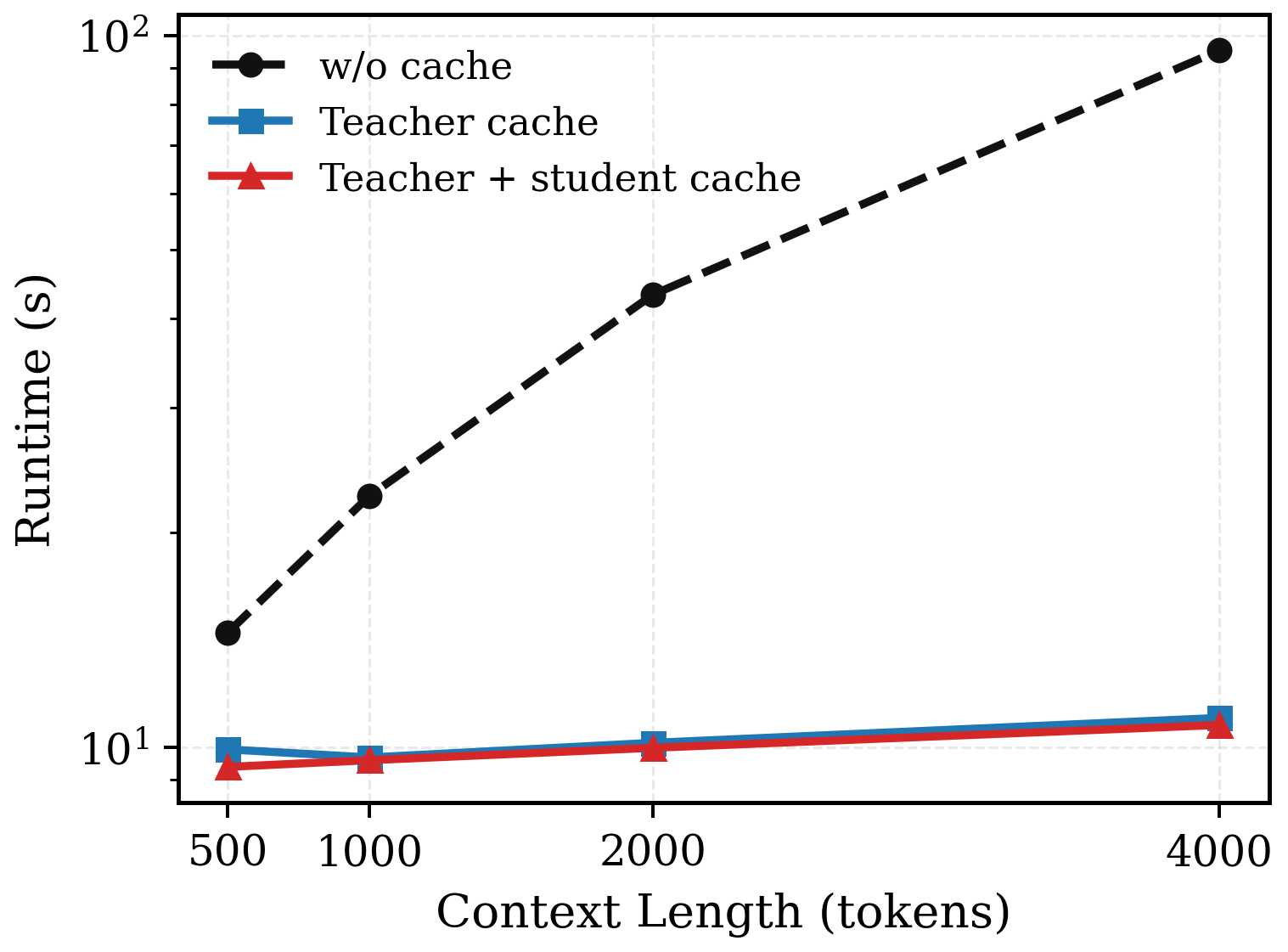}
        \vspace{-20pt}
        \caption{Runtime Comparison}
        % \label{fig:fig2}
    \end{subfigure}
    \hfill
    \begin{subfigure}[t]{0.32\linewidth}
        \centering
        \includegraphics[width=\linewidth]{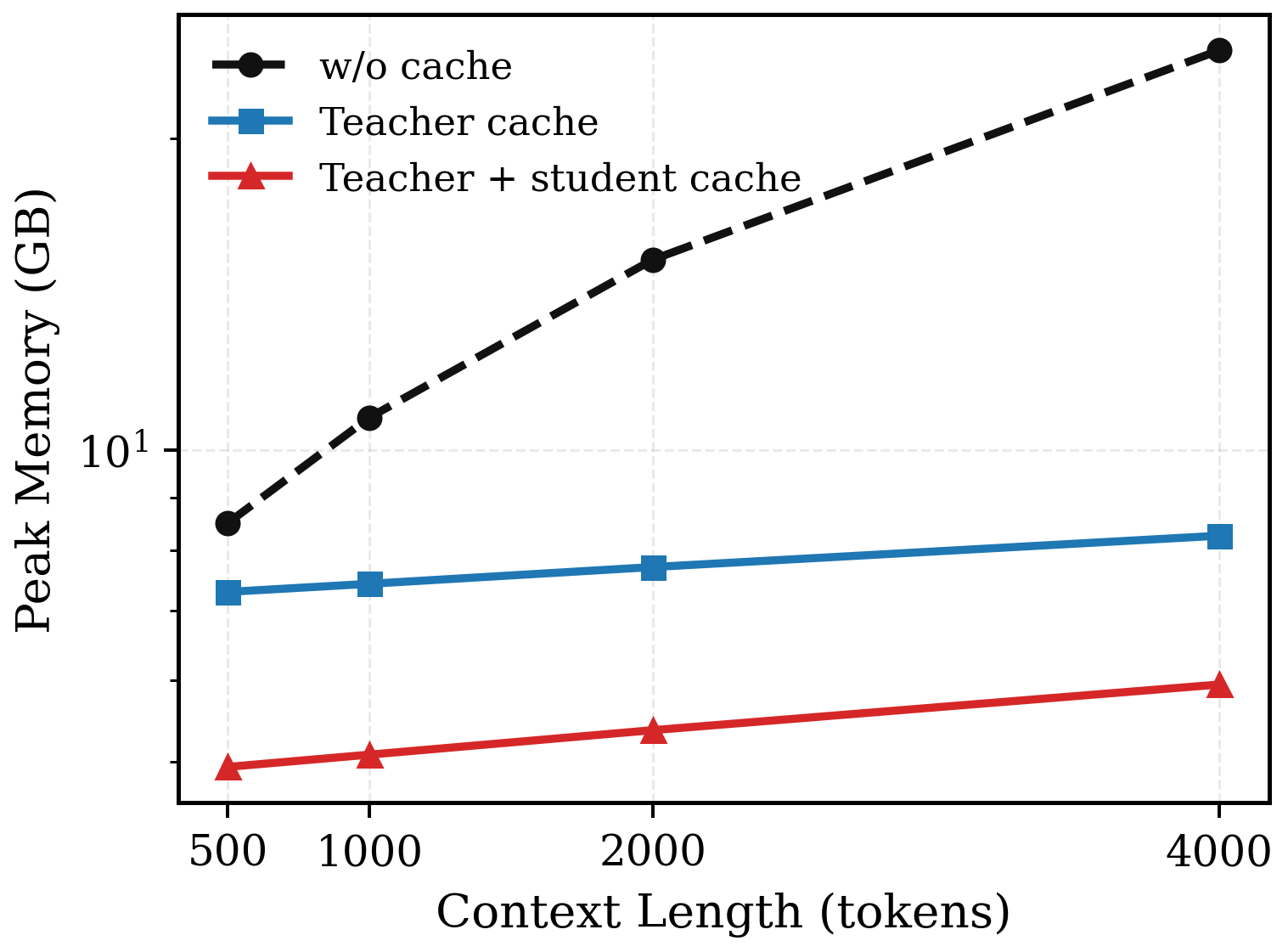}
        \vspace{-20pt}
        \caption{Memory Comparison}
        % \label{fig:fig3}
    \end{subfigure}
    \vspace{-8pt}
    \caption{Training efficiency comparison under varying context lengths.}
    \label{fig:training_efficiency}
    \vspace{-15pt}
\end{figure*}

\subsection{Whether to Activate}
\label{sec:activate}
The results in Section~\ref{sec:exp_retrieve} show that retrieved adapters can serve as effective latent memories for context-specific queries. 
However, practical systems must also decide \emph{whether} to activate a retrieved memory, since irrelevant adapters may interfere with the base model's general knowledge on context-agnostic queries. 
We evaluate this issue in a hybrid setting that combines context-specific queries from NarrativeQA and context-agnostic queries from CommonsenseQA, where Self-Gating decides whether to activate the adapter.
For isolating activation control, Table~\ref{tab:gating_results} evaluates Self-Gating on the top-1 retrieved adapter. This isolates whether the system can suppress an irrelevant latent memory, independent of the internal routing.

As shown in Table~\ref{tab:gating_results}, always activating an adapter substantially degrades CommonsenseQA performance. 
Cumulative methods using the \textit{latest} adapter suffer large drops, while our retrieval-only variant reduces CommonsenseQA accuracy from 47.80\% to 37.30\% on \textsc{Qwen2.5-0.5B}, and from 85.18\% to 72.40\% on \textsc{Qwen2.5-7B}. 
These results indicate that retrieval alone is insufficient: retrieved latent memories must still be selectively activated.
With gating, our method largely mitigates this interference. 
CommonsenseQA accuracy improves to 45.62\% on \textsc{Qwen2.5-0.5B} and 79.44\% on \textsc{Qwen2.5-7B}, while maintaining competitive NarrativeQA performance. 
These results show that our proposed Self-Gating mechanism enables the model to use latent memory only when necessary and fall back to the base model otherwise.

\begin{figure}[t]
    \centering
    \includegraphics[width=0.9\linewidth]{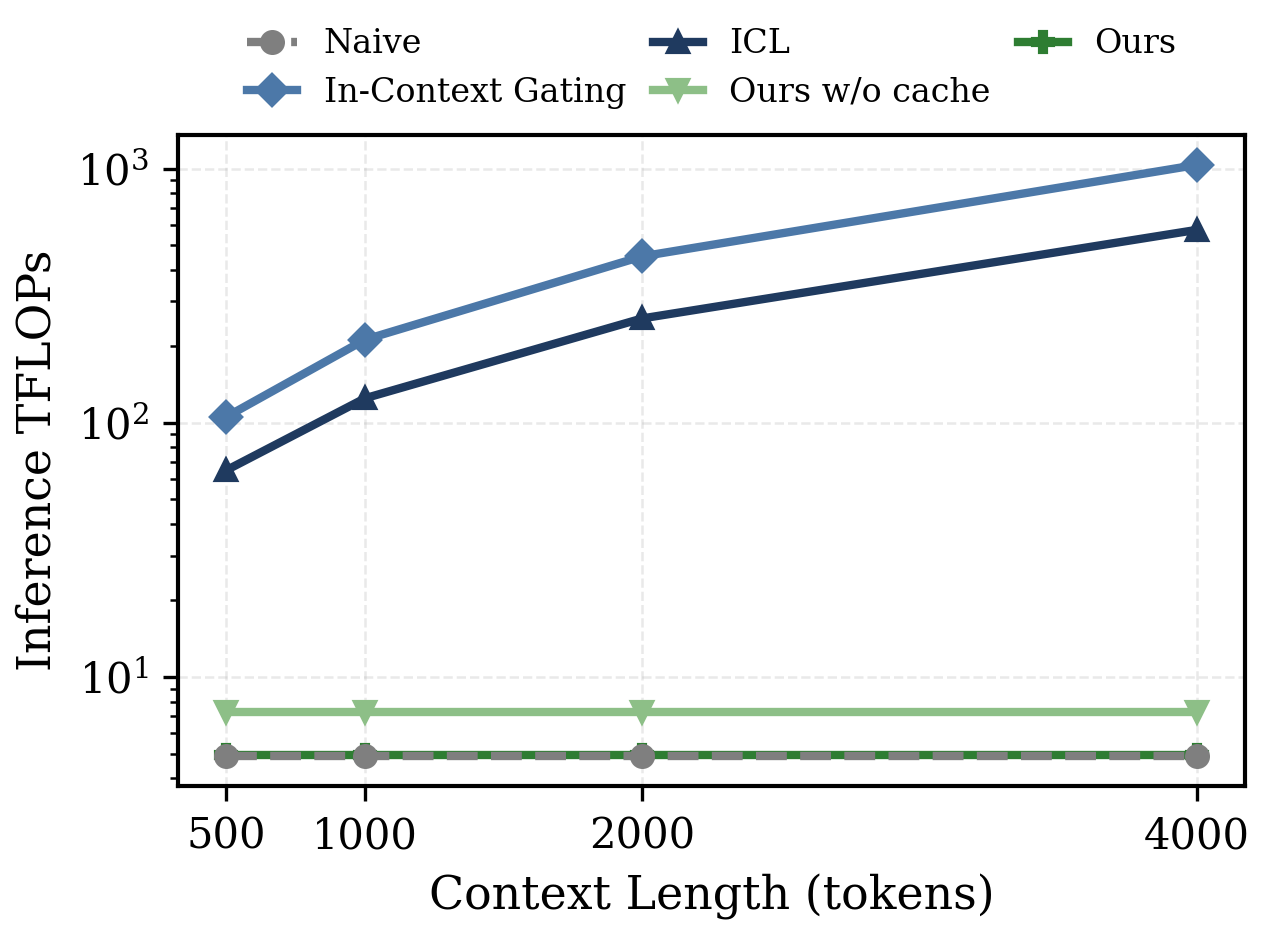}
    \vspace{-12pt}
    \caption{Inference efficiency comparison for gating.}
    \label{fig:gating_efficiency}
    \vspace{-15pt}
\end{figure}

\vspace{-5pt}
\paragraph{Efficiency}

Figure~\ref{fig:gating_efficiency} compares the inference cost of different activation strategies. 
Following the previous setting, \textit{In-Context Gating} denotes gating based on the first token generated with the retrieved context.
Unlike in-context methods, whose cost grows with the context length, our method activates memories by selectively enabling LoRA adapters and reusing the shared prefix cache.
At 4000 tokens, it reduces FLOPs by $210.0\times$ over in-context gating and $116.6\times$ over ICL, while introducing only about $1\%$ overhead over naive generation.
\vspace{-5pt}
\subsection{Training Efficiency}
While CD offers an effective way to reduce inference cost, its training cost increases with the context length~\citep{charakorn2026doc2lora,wang2024templora, cao2025infiniteicl}.
We evaluate training efficiency by varying the context length from 500 to 4000 tokens.
As shown in Figure~\ref{fig:training_efficiency}, compared with the w/o cache variant that repeatedly recomputes the full context, teacher-side cache reuse and student-side prefix caching substantially reduce computation.
At the longest context length, the full caching pipeline achieves an $8.4\times$ speedup, a $4.1\times$ reduction in peak memory usage, and a $21.9\times$ reduction in FLOPs.
Moreover, whereas the baseline cost grows rapidly with context length, our cached pipeline keeps training time nearly constant, indicating that the cost of long-context processing is effectively amortized.
Note that while off-policy methods~\citep{eyuboglu2025cartridges, wang2024templora, cao2025infiniteicl, snell2022learning} only need to generate labels for a single turn, the on-policy methods~\citep{ye2026OPCD, zhang2026opsdl} roll out responses online and compute logits conditioned on the context, making teacher-side cache reuse even more critical.

\subsection{Generalization and Ablation}
\label{sec:generalization}
We further investigate the generality of the proposed method. Table~\ref{tab:cache_comparison} reports the performance of different distillation variants with and without caching, with detailed distillation settings provided in Section~\ref{apdx:exp_distill}. 
Overall, the results suggest that the proposed cache-sharing is not specific to any single distillation objective. Across NarrativeQA and SQuAD, cache-sharing distillation largely preserves the average performance of the corresponding non-caching variants, and in some cases even yields improvements under certain objectives.
These findings suggest that our proposed cache-sharing distillation can be seamlessly combined with diverse distillation algorithms while maintaining effectiveness relative to standard non-cached distillation.
We further conduct experiment on Llama3.1-8B~\citep{grattafiori2024llama} in Section~\ref{apdx:exp_base}. We also discuss the cache-sharing distillation, internal routing and Self-Gating threshold with empirical ablation study in Section~\ref{apdx:ablation}.

% \begin{table}[t]
% \centering
% \small
% \setlength{\tabcolsep}{3.5pt}
% \renewcommand{\arraystretch}{1.10}
% \caption{
% Comparison of performance on NarrativeQA and SQuAD with and without cache.
% }
% \vspace{-8pt}
% \label{tab:cache_comparison}
% \begin{tabular}{@{}lcccc@{}}
% \toprule
% \multirow{2}{*}{Method} & \multicolumn{2}{c}{NarrativeQA} & \multicolumn{2}{c}{SQuAD} \\
% \cmidrule(lr){2-3} \cmidrule(lr){4-5}
% & ROUGE-1 $\uparrow$ & ROUGE-L $\uparrow$ & EM $\uparrow$ & F1 $\uparrow$ \\
% \midrule

% Forward KL
% & 0.3269 & 0.3174 & 33.16 & 50.35 \\
% \variant{\textsc{ w/ Cache}}
% & 0.3123 & 0.3022 & 33.38 & 50.05 \\

% \addlinespace[2pt]
% Reverse KL
% & 0.3197 & 0.3101 & 33.91 & 51.60 \\
% \variant{\textsc{ w/ Cache}}
% & 0.3074 & 0.2973 & 34.10 & 51.16 \\

% \addlinespace[2pt]
% Top-$k$ Logits
% & 0.3078 & 0.2981 & 31.30 & 49.90 \\
% \variant{\textsc{ w/ Cache}}
% & 0.3018 & 0.2919 & 32.25 & 50.32 \\

% \addlinespace[2pt]
% EMA Teacher
% & 0.1618 & 0.1525 & 13.17 & 29.67 \\
% \variant{\textsc{ w/ Cache}}
% & 0.1857 & 0.1763 & 16.62 & 32.09 \\

% \midrule
% Average
% & \textbf{0.2791} & \textbf{0.2695} & 27.89 & 45.38 \\
% \variant{\textsc{ w/ Cache}}
% & 0.2768 & 0.2669 & \textbf{29.09} & \textbf{45.91} \\
% \bottomrule
% \end{tabular}
% \vspace{-15pt}
% \end{table}

\begin{table}[t]
\centering
\small
\setlength{\tabcolsep}{3.5pt}
\renewcommand{\arraystretch}{1.0}
\caption{
Comparison of performance on NarrativeQA and SQuAD with Qwen2.5-0.5B.
}
\vspace{-8pt}
\label{tab:cache_comparison}
\begin{tabular}{@{}lcccc@{}}
\toprule
\multirow{2}{*}{Method} & \multicolumn{2}{c}{NarrativeQA} & \multicolumn{2}{c}{SQuAD} \\
\cmidrule(lr){2-3} \cmidrule(lr){4-5}
& ROUGE-1 $\uparrow$ & ROUGE-L $\uparrow$ & EM $\uparrow$ & F1 $\uparrow$ \\
\midrule

Forward KL
& 32.69 & 31.74 & 33.16 & 50.35 \\
\variant{\textsc{ w/ Cache}}
& 31.23 & 30.22 & 33.38 & 50.05 \\

\addlinespace[2pt]
Reverse KL
& 31.97 & 31.01 & 33.91 & 51.60 \\
\variant{\textsc{ w/ Cache}}
& 30.74 & 29.73 & 34.10 & 51.16 \\

\addlinespace[2pt]
Top-$k$ Logits
& 30.78 & 29.81 & 31.30 & 49.90 \\
\variant{\textsc{ w/ Cache}}
& 30.18 & 29.19 & 32.25 & 50.32 \\

\addlinespace[2pt]
EMA Teacher
& 16.18 & 15.25 & 13.17 & 29.67 \\
\variant{\textsc{ w/ Cache}}
& 18.57 & 17.63 & 16.62 & 32.09 \\

\midrule
Average
& \textbf{27.91} & \textbf{26.95} & \underline{27.89} & \underline{45.38} \\
\variant{\textsc{ w/ Cache}}
& \underline{27.68} & \underline{26.69} & \textbf{29.09} & \textbf{45.91} \\
\bottomrule
\end{tabular}
\vspace{-15pt}
\end{table}
\vspace{-5pt}
\section{Conclusion}
\vspace{-5pt}
We formulate context distillation as latent memory management. Instead of accumulating all contexts into a single parameter state, we distill each context into an independent LoRA adapter and manage these adapters through retrieval and Self-Gating. This design enables modular storage, query-aware memory selection, and safe fallback to the base model when latent memory is unnecessary.
Experiments show that our method outperforms cumulative distillation baselines in realistic non-oracle settings and preserves general model capabilities on context-agnostic queries, with a lightweight management system.

% \section{Limitation}
% \todo{limitation todo}

\clearpage
\section*{Limitations}
Although context distillation with LoRA as latent memory can reduce inference cost, it currently still exhibits a performance gap compared with ICL under the oracle setting. Future work may close this gap by developing more effective distillation strategies, such as on-policy distillation methods~\citep{ye2026OPCD,zhang2026opsdl}.

Furthermore, storage overhead remains a limitation, as saving a separate adapter can still be costly. Future work may investigate more parameter-efficient modules for context distillation, such as BitFit~\citep{zaken2022bitfit}, to further reduce the memory footprint.

\bibliography{custom}

\clearpage
\appendix
\section{Cumulative Context Distillation}
\label{apdx:cumulative}
Given a stream of contextual data $\mathcal{C} = \{c_1, c_2, \dots, c_t\}$, existing methods such as InfiniteICL \citep{cao2025infiniteicl} and TempLora \citep{wang2024templora} update the model's latent knowledge following a cumulative distillation paradigm. Specifically, at step $i$, the model parameters $\theta_{i-1}$ are updated to $\theta_i$ by minimizing the Kullback-Leibler (KL) divergence between the updated model and the previous model conditioned on the new context:
\begin{equation}
\theta_i = \mathop{\arg\min}_{\theta} D_{KL} \big( \pi_{\theta_{i-1}}(\cdot|q, c_i) \parallel \pi_{\theta}(\cdot|q) \big)
% \label{eq:cdistillation}
\end{equation}
where $q$ is the query.

\begin{figure}
    \centering
    \includegraphics[width=1.0\linewidth]{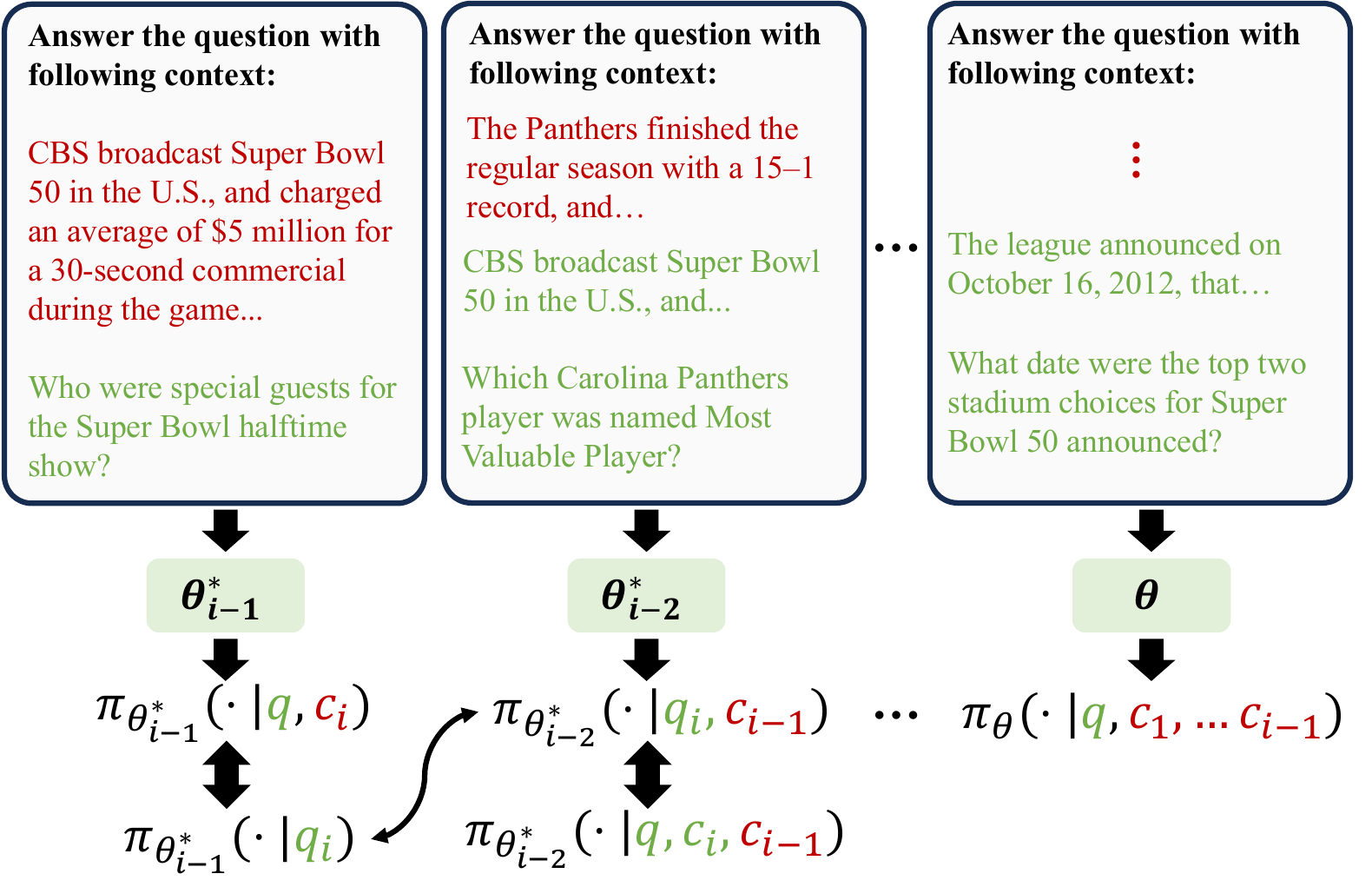}
    \caption{Theoretical upper bound of cumulative distillation paradigm.}
    \label{fig:upper_bound}
\end{figure}

To understand the theoretical limit of this paradigm, let us consider an ideal scenario. Assume that for all steps $i$, there exists an optimal set of parameters $\theta_{i}^*$ such that  the unconditioned output distribution perfectly matches the context-conditioned distribution from the previous step:
\begin{equation}
    \pi_{\theta_{i}^*}(y|q) = \pi_{\theta_{i-1}^*}(y|q, c_i), \forall q
\label{eq:ideal_step}
\end{equation}
with the initial state defined as:
\begin{equation}
    \pi_{\theta_{1}^*}(y|q) = \pi_{\theta}(y|q, c_1), \forall q
\label{eq:ideal_init}
\end{equation}
where $\theta$ denotes the parameters of the base model. Note that prepending context $c_i$ to query $q$ can be conceptually viewed as forming an augmented query $q_i = [c_i, q]$. As illustrated in Figure~\ref{fig:upper_bound}, by recursively applying Eq. \ref{eq:ideal_step} and Eq. \ref{eq:ideal_init}, we can derive the equivalent distribution at step $i$:
\begin{equation}
    \pi_{\theta_{i}^*}(y|q) = \pi_{\theta}(y|q, [c_1, c_2, \dots, c_i]), \forall q
\label{eq:upper_bound}
\end{equation}
Equation \ref{eq:upper_bound} reveals a critical property: 
under this idealized recursive teacher assumption, the target distribution coincides with the base model conditioned on the full concatenated context.

Consequently, this cumulative paradigm suffers from several inherent limitations:
\begin{itemize}
    \item \textbf{Inherited Long-Context Degradation:} Since the theoretical limit relies on the base model processing $[c_1, \dots, c_i]$, the distilled model may inherit the base model's weaknesses regarding long context.
    \item \textbf{Noise Assimilation:} In real-world streaming data, not all contexts are beneficial for answering subsequent queries. The standard cumulative approach indiscriminately distills all provided tokens, forcing the model to absorb noise and irrelevant information into its latent knowledge.
\end{itemize}

\section{Dataset}
\label{apdx:dataset}
\begin{table*}[h]
\centering
% \small
\setlength{\tabcolsep}{5pt}
\caption{Dataset statistics for NarrativeQA~\citep{kovcisky2018narrativeqa} and SQuAD~\citep{rajpurkar2016squad}. Document length, queries per document, and query length are reported as mean $\pm$ standard deviation; all lengths are measured in tokens.}
\begin{tabular}{llcccc}
\toprule
Dataset & Split & \#Docs & Doc Length & \#Queries per Doc & Query Length \\
\midrule
\multirow{2}{*}{NarrativeQA}
& Train     & \multirow{2}{*}{355} & \multirow{2}{*}{$731.44 \pm 259.76$} & $1000.00 \pm 0.00$ & $13.77 \pm 6.88$ \\
& Test      &                       &                                       & $29.74 \pm 2.45$   & $10.92 \pm 3.77$ \\
\midrule
\multirow{2}{*}{SQuAD}
& Train     & \multirow{2}{*}{300} & \multirow{2}{*}{$158.93 \pm 80.08$}  & $1000.00 \pm 0.00$ & $13.54 \pm 5.42$ \\
& Test &                       &                                       & $8.81 \pm 4.27$    & $12.71 \pm 4.15$ \\
\bottomrule
\end{tabular}

\label{tab:dataset_stats}
\end{table*}

\subsection{Synthetic Query Generation}
\label{apdx:query-generation}

\begin{promptbox}[label={prompt:query-generation}]{Query Generation}

Read the following story context and generate \placeholder{batch\_size}
distinct questions based on it.

\textbf{Requirements:}
\begin{enumerate}
    \item The questions must be highly related to the context.
    \item Output the questions as a numbered list, e.g., ``1. Question ... 2. Question ...''.
    \item Do not provide answers, options, or any other text. Only the questions.
\end{enumerate}

\textbf{Context:}

\placeholder{context}

\textbf{Questions:}

\end{promptbox}

Following previous works~\citep{cao2025infiniteicl,eyuboglu2025cartridges}, we construct synthetic queries from the NarrativeQA and SQuAD corpus.
For each document, we use the document summary as the source
context and prompt an instruction-tuned causal language model to generate
natural-language questions. The prompt requires the model to produce a
numbered list of distinct questions that are highly related to the given
context, while explicitly excluding answers, answer options, or additional
explanatory text. The prompt template is shown in
Prompt~\ref{prompt:query-generation}.

For each document, we repeatedly sample from the generator until collecting
up to $k=1000$ valid questions. We use stochastic decoding with temperature
sampling to encourage diversity. Since the raw model outputs may contain
formatting artifacts, we parse the generations with a regular expression that
extracts interrogative sentences from numbered or bulleted lists. We then
normalize each candidate by removing assistant-style prefixes and truncating
the text at the first question mark.

% To improve the semantic grounding of the generated queries, we further apply
% an embedding-based filtering step. Specifically, we encode each candidate
% query using Qwen3-Embedding-0.6B and compare it with the pre-computed embedding
% of the corresponding document summary. Candidate queries whose similarity
% score is lower than a threshold $\lambda=0.4$ are discarded. The remaining
% questions are paired with their document identifiers and used as synthetic
% queries in our downstream experiments.

% \begin{figure}[t]
% \centering
% \fbox{
% \begin{minipage}{0.92\linewidth}
% \small
% \textbf{Prompt for query generation}

% \vspace{0.5em}

% Read the following story context and generate \texttt{\{batch\_size\}}
% distinct questions based on it.

% \vspace{0.5em}

% \textbf{Requirements:}
% \begin{enumerate}
%     \item The questions must be highly related to the context.
%     \item Output the questions as a numbered list, e.g., ``1. Question ... 2. Question ...''.
%     \item Do not provide answers, options, or any other text. Only the questions.
% \end{enumerate}

% \textbf{Context:}

% \texttt{\{context\}}

% \vspace{0.5em}

% \textbf{Questions:}
% \end{minipage}
% }
% \caption{Prompt template used for generating synthetic queries from NarrativeQA document summaries.}
% \label{fig:query-generation-prompt}
% \end{figure}

\subsection{Dataset Statistics}
We report the dataset statistics used in our paper in Table ~\ref{tab:dataset_stats}. We use the document from the test dataset of NarrativeQA~\citep{kovcisky2018narrativeqa} and validation dataset of SQuAD~\citep{rajpurkar2016squad}. For train split, we generate queries with the process in Section~\ref{apdx:query-generation}, and generate corresponding responses with base model during training. For test split, we use the original queries and answers in the test dataset of NarrativeQA and validation dataset of SQuAD.

\section{Experiment Details}
We evaluate in a document-level adaptation setting. At distillation time, the system has access to the target document but not to the benchmark evaluation questions or gold answers. We generate synthetic training queries from the document and teacher responses from the base model, then evaluate on the original benchmark questions. Thus, our setting measures whether a document can be compiled into a reusable latent memory for future queries about that document.
\begin{table}[t]
\centering
\small
\setlength{\tabcolsep}{5.2pt}
\renewcommand{\arraystretch}{1.14}
\begin{threeparttable}
\caption{Training hyperparameters for LoRA distillation.}
\label{tab:training_hparams}

\begin{tabular}{@{}p{0.42\linewidth}p{0.52\linewidth}@{}}
\toprule
\textbf{Hyperparameter} & \textbf{Value} \\
\midrule

\rowcolor{gray!10}
\multicolumn{2}{@{}l}{\textbf{LoRA Configuration}} \\
\addlinespace[2pt]
Rank $r$ & 64 \\
Scaling factor $\alpha$ & 128 \\
Dropout & 0.05 \\
Target modules & Query, key, value, output, gate, up, and down projections \\

\addlinespace[3pt]
\rowcolor{gray!10}
\multicolumn{2}{@{}l}{\textbf{Optimization}} \\
\addlinespace[2pt]
Learning rate & $5 \times 10^{-4}$ \\
Training epochs & 10 \\
Training batch size & 16 \\
Learning rate scheduler & Cosine \\

\addlinespace[3pt]
\rowcolor{gray!10}
\multicolumn{2}{@{}l}{\textbf{Generation}} \\
\addlinespace[2pt]
Max new tokens & 50 \\

\bottomrule
\end{tabular}
\end{threeparttable}
\vspace{-6pt}
\end{table}
\label{adpx:exp_detail}
\subsection{Hyperparameter}
For all context distillation method in our paper, we use the same hyperparameter for training, as shown in table~\ref{tab:training_hparams}. 
% Results are reported with a single generation.
\subsection{Environment}
All experiments were conducted on a single GPU. For Qwen2.5-0.5B, we used an NVIDIA L40 GPU, while experiments with Qwen2.5-7B and Llama-3.1-8B were run on an NVIDIA H100 GPU. Each training run was completed within approximately 1--2 days.
\subsection{Metrics}
We evaluate SQuAD with Exact Match (EM) and token-level F1, following the standard SQuAD evaluation protocol. Both predictions and reference answers are normalized prior to scoring by lowercasing, removing punctuation and English articles, and fixing whitespace. For instances with multiple reference answers, we compute the score against each reference and report the maximum score.

For NarrativeQA, we report ROUGE-1 and ROUGE-L, computed using the HuggingFace \texttt{evaluate} implementation of ROUGE with its default configuration. Model predictions are compared against the provided gold reference answers.

% \subsection{Baseline Evaluation}
% \label{apdx:baseline_eval}
% \paragraph{Lora-only}
% For lora-only method, all memories are saved within the lora. Therefore, given a query $q^i$ from document $c_i$ with the latent memory $\Delta \theta_{i}$:
% \begin{itemize}
%     \item \textit{Latest} setting uses the latest adapter $\Delta \theta_n$, i.e. generate response with $f_{\theta+\Delta \theta_n}(q_i)$
%     \item \textit{Oracle} settings assume access to the correct adapter $\Delta \theta_{i}$, i.e. generate response with $f_{\theta+\Delta \theta_i}(q_i)$
%     \item \textit{Shift} setting evaluates a query $q^i$ from document $c_i$ with the next adapter $\Delta \theta_{i+1}$, to test the forgetting problem, i.e. generate response with $f_{\theta+\Delta \theta_{i+1}}(q_i)$
% \end{itemize}

% \paragraph{Lora+Context}
% InfiniteICL~\citep{cao2025infiniteicl} takes adapter as long term memory and context as short term memory. Therefore, given a query $q^i$ from document $c_i$ with the latent memory $\Delta \theta_{i}$:
% \begin{itemize}
%     \item \textit{Latest}: generate response with $f_{\theta+\Delta \theta_{n-1}}(q_i,c_n)$
%     \item \textit{Oracle}: generate response with $f_{\theta+\Delta \theta_{i-1}}(q_i,c_i)$
%     \item \textit{Shift}: generate response with $f_{\theta+\Delta \theta_{i}}(q_i,c_{i+1})$
% \end{itemize}

\subsection{Baseline Evaluation}
\label{apdx:baseline_eval}

\paragraph{LoRA-only.}
In the LoRA-only setting, all document-specific information is stored in the LoRA adapter. Given a query $q_i$ associated with document $c_i$, whose corresponding latent memory is encoded by the adapter $\Delta\theta_i$, we evaluate the following settings:
\begin{itemize}
    \item \textit{\textbf{Latest}}: uses the most recent adapter $\Delta\theta_n$ to answer the query, i.e., generates the response as $f_{\theta+\Delta\theta_n}(q_i)$.
    \item \textit{\textbf{Oracle}}: assumes access to the correct adapter $\Delta\theta_i$ associated with document $c_i$, i.e., generates the response as $f_{\theta+\Delta\theta_i}(q_i)$.
    \item \textit{\textbf{Shift}}: evaluates the query $q_i$ using the next adapter $\Delta\theta_{i+1}$, i.e., generates the response as $f_{\theta+\Delta\theta_{i+1}}(q_i)$. This setting is designed to probe whether information from $c_i$ is forgotten or overwritten after updating to the subsequent adapter.
\end{itemize}

\paragraph{LoRA+Context.}
InfiniteICL~\citep{cao2025infiniteicl} treats the LoRA adapter as long-term memory and the input context as short-term memory. Given a query $q_i$ associated with document $c_i$, with the corresponding latent memory encoded by $\Delta\theta_i$, we evaluate:
\begin{itemize}
    \item \textit{\textbf{Latest}}: uses the latest available context $c_n$ together with the preceding adapter $\Delta\theta_{n-1}$, i.e., generates the response as $f_{\theta+\Delta\theta_{n-1}}(q_i, c_n)$.
    \item \textit{\textbf{Oracle}}: assumes access to the correct context $c_i$ and the adapter immediately preceding it, $\Delta\theta_{i-1}$, i.e., generates the response as $f_{\theta+\Delta\theta_{i-1}}(q_i, c_i)$.
    \item \textit{\textbf{Shift}}: evaluates the query $q_i$ with the subsequent context $c_{i+1}$ and the adapter $\Delta\theta_i$, i.e., generates the response as $f_{\theta+\Delta\theta_i}(q_i, c_{i+1})$.
\end{itemize}
\subsection{Router Training Details}
\label{apdx:router-training}

We train a lightweight router to select one adapter from the retrieved top-$k$ candidates for each query. Given a query $q$, retrieval returns
$\mathcal{C}_k(q)=\{c_1,\ldots,c_k\}$ with similarity scores $s(q,c_i)$. A separate router is trained for each dataset, routing mode, and value of $k$.

For each candidate $c_i$, we use the feature vector
\begin{equation}
\mathbf{x}_{q,i}
=
\left[
s(q,c_i),\;
s(q,c_1)-s(q,c_i),\;
H_{q,i}^{(1)},\;
\mathbf{h}_{q,i}^{(1)}
\right],
\end{equation}
where $s(q,c_i)$ is the retrieval similarity, $s(q,c_1)-s(q,c_i)$ is the gap to the top-ranked candidate, $H_{q,i}^{(1)}$ is the entropy of the first-token predictive distribution, and $\mathbf{h}_{q,i}^{(1)}$ denotes the first 128 dimensions of the first-token hidden state.

We use candidate-level binary supervision:
\begin{equation}
y_{q,i}=\mathbbm{1}[c_i=c^\star(q)],
\end{equation}
where $c^\star(q)$ is the gold document/adapter id. If $c^\star(q)\notin\mathcal{C}_k(q)$, all candidates for $q$ are labeled as negative. The router is therefore trained as a binary scorer.
All candidate features from the training split are flattened and standardized using training-set statistics.
The router is a two-layer MLP that outputs one scalar logit per candidate:
\begin{equation}
r_\theta(\mathbf{x})=\mathrm{MLP}_\theta(\tilde{\mathbf{x}}).
\end{equation}

We train the router with weighted binary cross-entropy loss, using
\begin{equation}
w_+ = \frac{N_{\mathrm{neg}}}{\max(N_{\mathrm{pos}},1)}
\end{equation}
to compensate for class imbalance, and optimize with AdamW. Unless otherwise specified, we use 100 epochs, learning rate $5\times10^{-2}$, weight decay $10^{-3}$, and full-batch training.

At inference time, the router scores candidates independently and selects
\begin{equation}
\hat{c}(q)
=
\arg\max_{c_i\in\mathcal{C}_k(q)}
r_\theta(\mathbf{x}_{q,i}),
\end{equation}
which is then used as the adapter id for downstream LoRA generation.

The router is trained only on synthetic QA examples in Section~\ref{apdx:query-generation}. For each dataset, we first construct a training split from generated questions associated with the adapter pool, and use the corresponding document id as the positive adapter label. In contrast, evaluation is performed on the original real QA split: SQuAD uses real questions from the SQuAD validation split, and NarrativeQA uses real questions from the NarrativeQA test split.

\subsection{Gating Threshold Selection}
\label{adpx:gating-threshold}

We use the entropy of the first generated token as the gating signal. Given a query--adapter pair, the LoRA-adapted model produces the next-token logits at the first decoding step. We convert the logits into a probability distribution
\[
p_0(v) = \mathrm{softmax}(z_0)_v,
\]
and compute the first-token entropy
\[
\mathcal{H}_0 = - \sum_{v \in \mathcal{V}} p_0(v) \log p_0(v).
\]
A lower value of $\mathcal{H}_0$ indicates that the LoRA adapter is more confident for the current query, while a higher value suggests that the query is out of distribution for the selected adapter. Therefore, we accept the LoRA adapter only when
\[
\mathcal{H}_0 < \lambda,
\]
and otherwise falls back to the base model.

The threshold $\lambda$ is selected on a held-out threshold-selection split, not on the final test split. Specifically, we use synthetic queries in Section~\ref{apdx:query-generation} for NarrativeQA and use the queries in CommonSenseQA training split. In contrast, evaluation is performed on real questions from the NarrativeQA test split and CommonSenseQA validation split.
We further analyze the sensitivity to $\lambda$ in Section~\ref{apdx:ablation_gating_thre}.
% In our paper, we use $\lambda=4.2$ for Qwen2.5-0.5B and $\lambda=4.0$ for Qwen2.5-7B.

% \section{Experiment with Llama}
% \label{apdx:exp_base}
% TODO

% \begin{figure*}[h]
%     \centering
%     \begin{subfigure}[t]{0.32\linewidth}
%         \centering
%         \includegraphics[width=\linewidth]{fig/training/paper_flops_total_absolute.png}
%         \vspace{-20pt}
%         \caption{Flops Comparison}
%         \label{fig:fig1}
%     \end{subfigure}
%     \hfill
%     \begin{subfigure}[t]{0.32\linewidth}
%         \centering
%         \includegraphics[width=\linewidth]{fig/training/paper_runtime_absolute.png}
%         \vspace{-20pt}
%         \caption{Runtime Comparison}
%         \label{fig:fig2}
%     \end{subfigure}
%     \hfill
%     \begin{subfigure}[t]{0.32\linewidth}
%         \centering
%         \includegraphics[width=\linewidth]{fig/training/paper_memory_absolute.png}
%         \vspace{-20pt}
%         \caption{Memory Comparison}
%         \label{fig:fig3}
%     \end{subfigure}
%     \vspace{-8pt}
%     \caption{Training efficiency}
%     \label{fig:three_figures}
%     \vspace{-15pt}
% \end{figure*}

\section{Integration with Different Distillation Objectives}
\label{apdx:exp_distill}

% \begin{table}[t]
% \centering
% \small
% \setlength{\tabcolsep}{3.5pt}
% \renewcommand{\arraystretch}{1.10}
% \caption{
% Comparison of performance on NarrativeQA and SQuAD with and without prefix caching.
% }
% \label{tab:cache_comparison}
% \begin{tabular}{@{}lcccc@{}}
% \toprule
% \multirow{2}{*}{Method} & \multicolumn{2}{c}{NarrativeQA} & \multicolumn{2}{c}{SQuAD} \\
% \cmidrule(lr){2-3} \cmidrule(lr){4-5}
% & ROUGE-1 $\uparrow$ & ROUGE-L $\uparrow$ & EM $\uparrow$ & F1 $\uparrow$ \\
% \midrule

% Forward KL
% & 0.3269 & 0.3174 & 33.16 & 50.35 \\
% \variant{\textsc{ w/ Cache}}
% & 0.3123 & 0.3022 & 33.38 & 50.05 \\

% \addlinespace[2pt]
% Reverse KL
% & 0.3197 & 0.3101 & 33.91 & 51.60 \\
% \variant{\textsc{ w/ Cache}}
% & 0.3074 & 0.2973 & 34.10 & 51.16 \\

% \addlinespace[2pt]
% Top-$k$ Logits
% & 0.3078 & 0.2981 & 31.30 & 49.90 \\
% \variant{\textsc{ w/ Cache}}
% & 0.3018 & 0.2919 & 32.25 & 50.32 \\

% \addlinespace[2pt]
% EMA Teacher
% & 0.1618 & 0.1525 & 13.17 & 29.67 \\
% \variant{\textsc{ w/ Cache}}
% & 0.1857 & 0.1763 & 16.62 & 32.09 \\

% \midrule
% Average
% & \textbf{0.2791} & \textbf{0.2695} & 27.89 & 45.38 \\
% \variant{\textsc{ w/ Cache}}
% & 0.2768 & 0.2669 & \textbf{29.09} & \textbf{45.91} \\
% \bottomrule
% \end{tabular}
% \vspace{-15pt}
% \end{table}

While the preceding experiments primarily adopt standard context distillation with forward KL divergence, our training pipeline is compatible with a broader class of distillation objectives. 
We therefore integrate the proposed prefix caching mechanism with several representative distillation algorithms and examine whether caching affects downstream task performance.
Specifically, we consider four commonly used distillation objectives.

\paragraph{Forward KL.}

Forward KL is widely adopted in prior context distillation methods~\citep{cao2025infiniteicl,eyuboglu2025cartridges,li2026lcc}. 
It encourages the student distribution without the context to match the teacher distribution conditioned on the full context:
\begin{equation}
    D_{\mathrm{KL}}\!\left(
    \pi_{\theta}(\cdot \mid q,c)
    \,\middle\|\,
    \pi_{\theta+\Delta\theta}(\cdot \mid q)
    \right).
\end{equation}

\paragraph{Reverse KL.}

Reverse KL has recently been used in on-policy distillation methods~\citep{shenfeld2026SDFT,hubotter2026SDPO,zhang2026opsdl}, where distillation is formulated from a reinforcement-learning perspective. 
The objective is given by
\begin{equation}
    D_{\mathrm{KL}}\!\left(
    \pi_{\theta+\Delta\theta}(\cdot \mid q)
    \,\middle\|\,
    \pi_{\theta}(\cdot \mid q,c)
    \right).
\end{equation}

\paragraph{Top-$k$ Logits.}

This objective applies KL divergence only over the top-$k$ tokens with the highest probabilities~\citep{raman2023distillation,ye2026OPCD}. 
Following OPCD~\citep{ye2026OPCD}, we use
\begin{equation}
    D_{\mathrm{KL}}^{\mathrm{top}\text{-}k}\!\left(
    \pi_{\theta+\Delta\theta}(\cdot \mid q)
    \,\middle\|\,
    \pi_{\theta}(\cdot \mid q,c)
    \right).
\end{equation}

\paragraph{EMA Teacher.}

In this setting, the teacher is not fixed during training. 
Instead, its parameters are updated as an exponential moving average of the student parameters:
\begin{equation}
    \bar{\theta}_{t}
    =
    \alpha \bar{\theta}_{t-1}
    +
    (1-\alpha)\theta_{t},
\end{equation}
where $\alpha \in [0,1)$ is the EMA decay rate. 
The distillation objective is then computed using the EMA teacher distribution:
\begin{equation}
    D_{\mathrm{KL}}\!\left(
    \pi_{\bar{\theta}_{t}}(\cdot \mid q,c)
    \,\middle\|\,
    \pi_{\theta_{t}}(\cdot \mid q)
    \right).
\end{equation}

\section{Efficiency }
\label{apdx:efficiency}

\subsection{training efficiency}

We evaluate training efficiency on a fixed NarrativeQA-sampled setting with
$N=1000$ question-answer pairs. The benchmark uses
Qwen2.5-0.5B with LoRA fine-tuning, bfloat16 precision, batch
size $16$, and $5$ warm-up steps. We evaluate
context lengths $C \in \{500,1000,2000,4000\}$ and normalize each answer to
$32$ tokens.

Each measured step performs a full optimizer update using a
temperature-scaled KL loss between teacher and student logits on the answer
tokens. Runtime includes both one-time KV precomputation and the measured
training loop:
\[
t_{\mathrm{total}} = t_{\mathrm{precompute}} + t_{\mathrm{train}}.
\]
Warm-up steps are excluded. Peak memory is measured with CUDA peak memory
statistics. FLOPs are estimated with DeepSpeed \texttt{FlopsProfiler} by
profiling one batch, scaling to 1000 queries.

\subsection{Inference efficiency}
We evaluate inference computation on NarrativeQA with Qwen2.5-0.5B under fixed-length generation. Each run uses
$100$ sampled queries and generates $16$ tokens per query. We report total
FLOPs. FLOPs are measured with DeepSpeed \texttt{FlopsProfiler}.
The reported total is
\[
\mathrm{FLOPs}_{\mathrm{total}}
=
\mathrm{FLOPs}_{\mathrm{routing}}
+
\mathrm{FLOPs}_{\mathrm{generation}},
\]
where cached variants decompose routing/generation into prefix-prefill and
cache-reuse terms.
\section{Ablation Study}
\label{apdx:ablation}

\begin{table}[t]
\centering
\small
\caption{Performance comparison under different retrieval settings on NarrativeQA. Best results within each model are highlighted in bold.}
\label{tab:qwen_rag_narrativeqa}
\setlength{\tabcolsep}{5pt}
\begin{tabular}{llccc}
\toprule
Model & Setting & Acc. & R-1 & R-L \\
\midrule
\multirow{3}{*}{\shortstack{Qwen 2.5\\0.5B}}
& w/o internal routing & 65.10\% & 23.47 & 22.54 \\
& Retrieval@3          & 67.30\% & \textbf{24.54} & \textbf{23.57} \\
& Retrieval@5          & \textbf{67.50\%} & 24.44 & 23.46 \\
\midrule
\multirow{3}{*}{\shortstack{Qwen 2.5\\7B}}
& w/o internal routing & 64.97\% & 27.18 & 26.41 \\
& Retrieval@3          & 67.59\% & 28.46 & 27.68 \\
& Retrieval@5          & \textbf{67.94\%} & \textbf{28.64} & \textbf{27.86} \\
\bottomrule
\end{tabular}
\end{table}

\begin{table}[t]
\centering
\small
\caption{Performance comparison under different retrieval settings on SQuAD. Best results within each model are highlighted in bold.}
\label{tab:qwen_rag_squad}
\setlength{\tabcolsep}{5pt}
\begin{tabular}{llccc}
\toprule
Model & Setting & Acc. & EM & F1 \\
\midrule
\multirow{3}{*}{\shortstack{Qwen 2.5\\0.5B}}
& w/o internal routing & 73.70\% & 26.73 & 41.26 \\
& Retrieval@3          & \textbf{76.90\%} & \textbf{28.30} & 43.11 \\
& Retrieval@5          & 76.60\% & 28.14 & \textbf{43.25} \\
\midrule
\multirow{3}{*}{\shortstack{Qwen 2.5\\7B}}
& w/o internal routing & 73.81\% & 34.85 & 44.94 \\
& Retrieval@3          & \textbf{76.91\%} & \textbf{36.34} & \textbf{46.57} \\
& Retrieval@5          & 76.00\% & 36.30 & 46.45 \\
\bottomrule
\end{tabular}
\end{table}
\subsection{Ablation on Internal Routing}

We ablate the effect of internal routing by varying the number of retrieved adapters available to the model. In this setting, \textit{w/o internal routing} serves as the baseline, where the model directly uses a single top-ranked retrieved adapter without further selection. In contrast, \textit{Retrieval@3} and \textit{Retrieval@5} enable internal routing, allowing the model to select an adapter from multiple retrieved candidates. We report both downstream task performance and routing accuracy, where the latter measures whether the routing module selects the appropriate adapter from the retrieved candidate set.

As shown in Tables~\ref{tab:qwen_rag_narrativeqa} and~\ref{tab:qwen_rag_squad}, enabling internal routing consistently improves performance across both model scales and datasets. For Qwen 2.5 0.5B on NarrativeQA, \textit{Retrieval@3} improves routing accuracy by 2.20, Rouge-1 by 1.07, and Rouge-L by 1.03 over the baseline without internal routing. \textit{Retrieval@5} further improves routing accuracy by 2.40 and achieves the best accuracy for this model, although its Rouge scores are slightly lower than those of \textit{Retrieval@3}. On SQuAD, the same model also benefits from internal routing: \textit{Retrieval@3} improves routing accuracy by 3.20, EM by 1.57, and F1 by 1.85. Increasing the retrieval size to 5 yields the largest F1 gain of 1.99, but leads to slightly smaller improvements in routing accuracy and EM compared with \textit{Retrieval@3}.

Similar trends are observed for Qwen 2.5 7B. On NarrativeQA, \textit{Retrieval@3} improves routing accuracy by 2.62, Rouge-1 by 1.28, and Rouge-L by 1.27. \textit{Retrieval@5} achieves the best overall performance, with gains of 2.97 in routing accuracy, 1.46 in Rouge-1, and 1.45 in Rouge-L. On SQuAD, \textit{Retrieval@3} obtains the best results, improving routing accuracy by 3.10, EM by 1.49, and F1 by 1.63. In contrast, \textit{Retrieval@5} brings smaller gains of 2.19 in routing accuracy, 1.45 in EM, and 1.51 in F1, suggesting that retrieving more adapters does not always lead to better routing decisions.

Overall, these results indicate that internal routing provides a clear benefit over relying on a single retrieved adapter. By routing among multiple retrieved adapters, the model can exploit complementary task-specific capabilities and produce more accurate predictions. The routing accuracy results further support this conclusion: enabling internal routing consistently improves adapter selection accuracy over the baseline, and improvements in routing accuracy generally align with gains in downstream performance. However, the comparison between \textit{Retrieval@3} and \textit{Retrieval@5} also shows that a larger candidate set may introduce additional ambiguity, especially on SQuAD, where \textit{Retrieval@3} already provides sufficient candidates for effective routing. This suggests that internal routing benefits from a balanced retrieval size that provides enough diversity while avoiding excessive noisy candidates.

\subsection{Ablation on Self-Gating Threshold}
\label{apdx:ablation_gating_thre}

We analyze how the first-token entropy threshold affects the activation behavior of Self-Gating in the hybrid setting. For each evaluation sample, we first retrieve the top-1 LoRA adapter using the same dense retriever as in the main hybrid evaluation. Given the retrieved adapter, we compute the first-token predictive entropy \(H_i\) under the LoRA-adapted model, using the same shared-prefix KV-cache computation as in the inference pipeline. Following the Self-Gating mechanism, the system activates the retrieved LoRA adapter when its first-token entropy is below a threshold \(\lambda\), and otherwise falls back to the base model:
\begin{equation}
g_i(\lambda)=\mathbf{1}\{H_i < \lambda\},
\end{equation}
where \(g_i(\lambda)=1\) denotes continuing generation with the retrieved LoRA adapter.
\begin{figure}[t]
    \centering
    \includegraphics[width=1.0\linewidth]{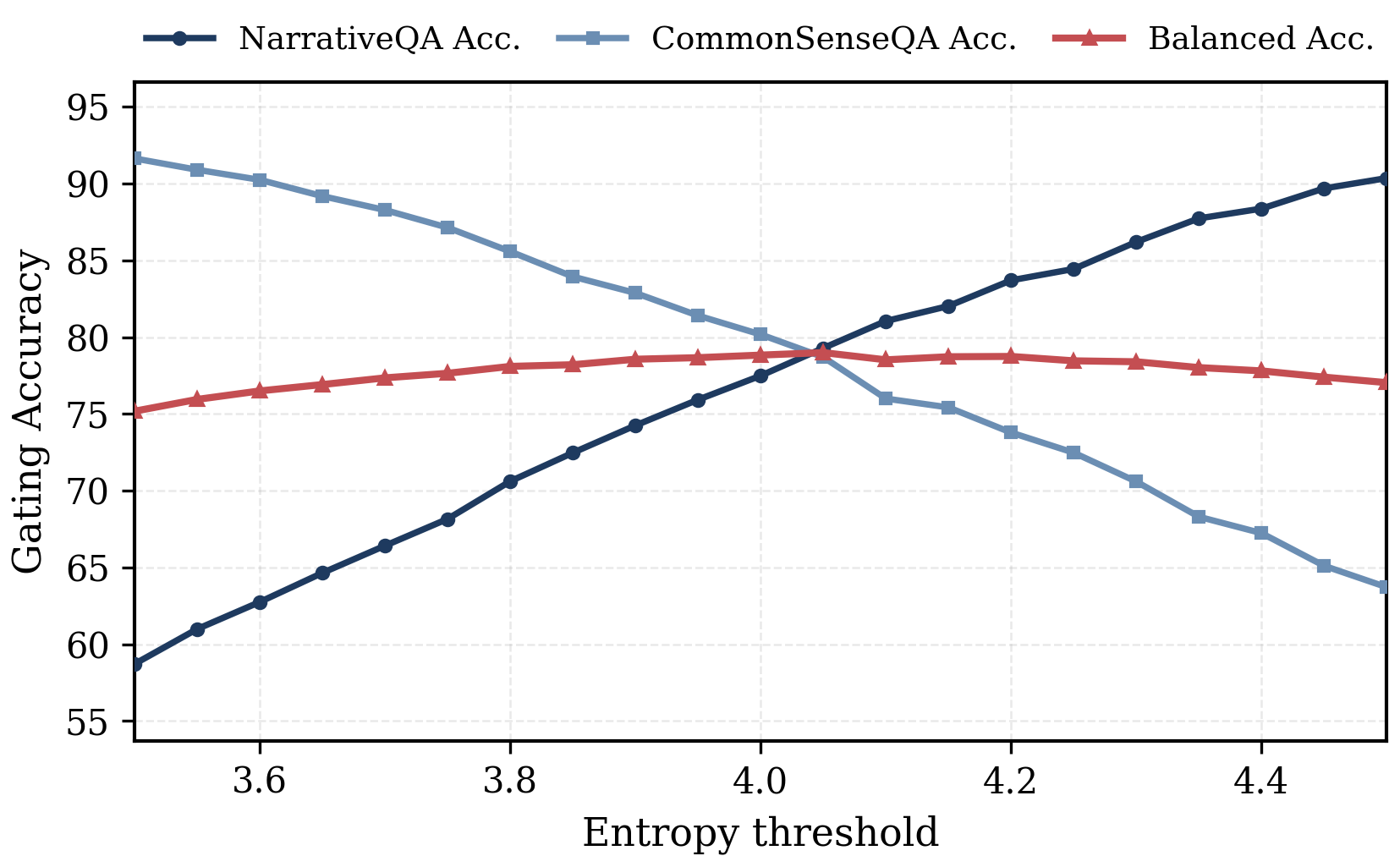}
    \caption{Effect of the first-token entropy threshold on Self-Gating. While the activation and fallback accuracies vary monotonically with the threshold, the balanced accuracy remains stable across a broad operating region, indicating that Self-Gating is insensitive to precise threshold tuning.}
    \label{fig:ablation_gating_threshold}
\end{figure}
In the hybrid evaluation, NarrativeQA queries are context-specific and are expected to benefit from the retrieved latent memory. Therefore, activating the retrieved LoRA adapter is treated as the correct gating decision for NarrativeQA. In contrast, CommonSenseQA queries are context-agnostic and should be answered using the base model without activating an irrelevant latent memory. Thus, falling back to the base model is treated as the correct gating decision for CommonSenseQA. The NarrativeQA gating accuracy is defined as
\begin{equation}
\mathrm{Acc}_{\mathrm{NQA}}(\lambda)
=
\frac{1}{|\mathcal{D}_{\mathrm{NQA}}|}
\sum_{i\in \mathcal{D}_{\mathrm{NQA}}}
\mathbf{1}\{H_i < \lambda\}.
\end{equation}

The CommonSenseQA gating accuracy is defined as
\begin{equation}
\mathrm{Acc}_{\mathrm{CSQA}}(\lambda)
=
\frac{1}{|\mathcal{D}_{\mathrm{CSQA}}|}
\sum_{i\in \mathcal{D}_{\mathrm{CSQA}}}
\mathbf{1}\{H_i \ge \lambda\}.
\end{equation}

We further report the balanced gating accuracy:
\begin{equation}
\mathrm{Acc}_{\mathrm{Balanced}}(\lambda)
=
\frac{1}{2}
\left(
\mathrm{Acc}_{\mathrm{NQA}}(\lambda)
+
\mathrm{Acc}_{\mathrm{CSQA}}(\lambda)
\right).
\end{equation}

We sweep the entropy threshold \(\lambda\) from \(3.5\) to \(4.5\) with a step size of \(0.05\), covering the operating region around \(\lambda=4.0\). Evaluation is conducted on the NarrativeQA test split and the CommonSenseQA validation split. Figure~\ref{fig:ablation_gating_threshold} plots the NarrativeQA activation accuracy, CommonSenseQA fallback accuracy, and balanced accuracy as functions of the entropy threshold.

The main observation is that Self-Gating is robust to the precise choice of the entropy threshold. Although changing \(\lambda\) shifts the relative preference between activating the retrieved LoRA adapter and falling back to the base model, the balanced accuracy remains nearly flat around the operating region. In particular, for thresholds from \(\lambda=3.75\) to \(\lambda=4.25\), the balanced accuracy stays within a narrow range from \(77.65\%\) to \(79.00\%\). The best value in this sweep is \(79.00\%\) at \(\lambda=4.05\), while neighboring thresholds achieve comparable performance.

This insensitivity indicates that Self-Gating does not rely on finely tuned thresholds to distinguish when latent memory should be activated. Instead, the first-token entropy of the LoRA-adapted model provides a stable confidence signal: across a broad interval of \(\lambda\), the system maintains a similar balance between using context-specific latent memory for NarrativeQA and abstaining from irrelevant memory activation for CommonSenseQA. These results support the practical robustness of the proposed gating mechanism in non-oracle hybrid settings.

\subsection{Ablation on Cache-Sharing Distillation}

\begin{table}[h]
\centering
\footnotesize
\setlength{\tabcolsep}{3pt}
\renewcommand{\arraystretch}{1.05}
\caption{Ablation study on Qwen2.5-0.5B evaluated on NarrativeQA and SQuAD.}
\label{tab:ablation_qwen}
\vspace{-6pt}
\begin{tabular}{@{}lcccc@{}}
\toprule
\multirow{2}{*}{Setting} 
& \multicolumn{2}{c}{NarrativeQA} 
& \multicolumn{2}{c}{SQuAD} \\
\cmidrule(lr){2-3} \cmidrule(lr){4-5}
& ROUGE-1 $\uparrow$ 
& ROUGE-L $\uparrow$ 
& EM $\uparrow$ 
& F1 $\uparrow$ \\
\midrule
Ours              & \textbf{31.23} & \textbf{30.22} & \textbf{33.38} & \textbf{50.05} \\
w/o cache-sharing & 26.99 & 26.00 & 27.02 & 43.47 \\
\bottomrule
\end{tabular}
\vspace{-10pt}
\end{table}

Previous work~\citep{wang2024templora} discuss the similar cache sharing mechanism for long text generation, that directly apply KV-Cache after LoRA updates, instead of recomputing them every time. Therefore, we conduct ablation study on whether cache-sharing distillation is important for the inference pipeline in our paper, i.e. we first compute KV-cache with base model, and reuse it with Lora model for generation.

As shown in Table~\ref{tab:ablation_qwen}, removing cache-sharing distillation consistently degrade performance on both datasets. On NarrativeQA, ROUGE-1 and ROUGE-L drop by 4.24 and 4.22, respectively. On SQuAD, EM and F1 decrease by 6.36 and 6.58. These results indicate that directly reusing KV-caches computed by the base model for the LoRA-adapted model introduces a notable mismatch.
This degradation is likely caused by the discrepancy between the base-model cached states and the updated attention computation after applying LoRA. Cache-sharing distillation explicitly exposes the model to this inference setting during training, encouraging the LoRA model to remain compatible with base-model KV-caches.

\section{Experiments Across Models}
\label{apdx:exp_base}

\begin{figure}[t]
    \centering

    \begin{subfigure}{1.0\linewidth}
        \centering
        \includegraphics[width=0.9\linewidth]{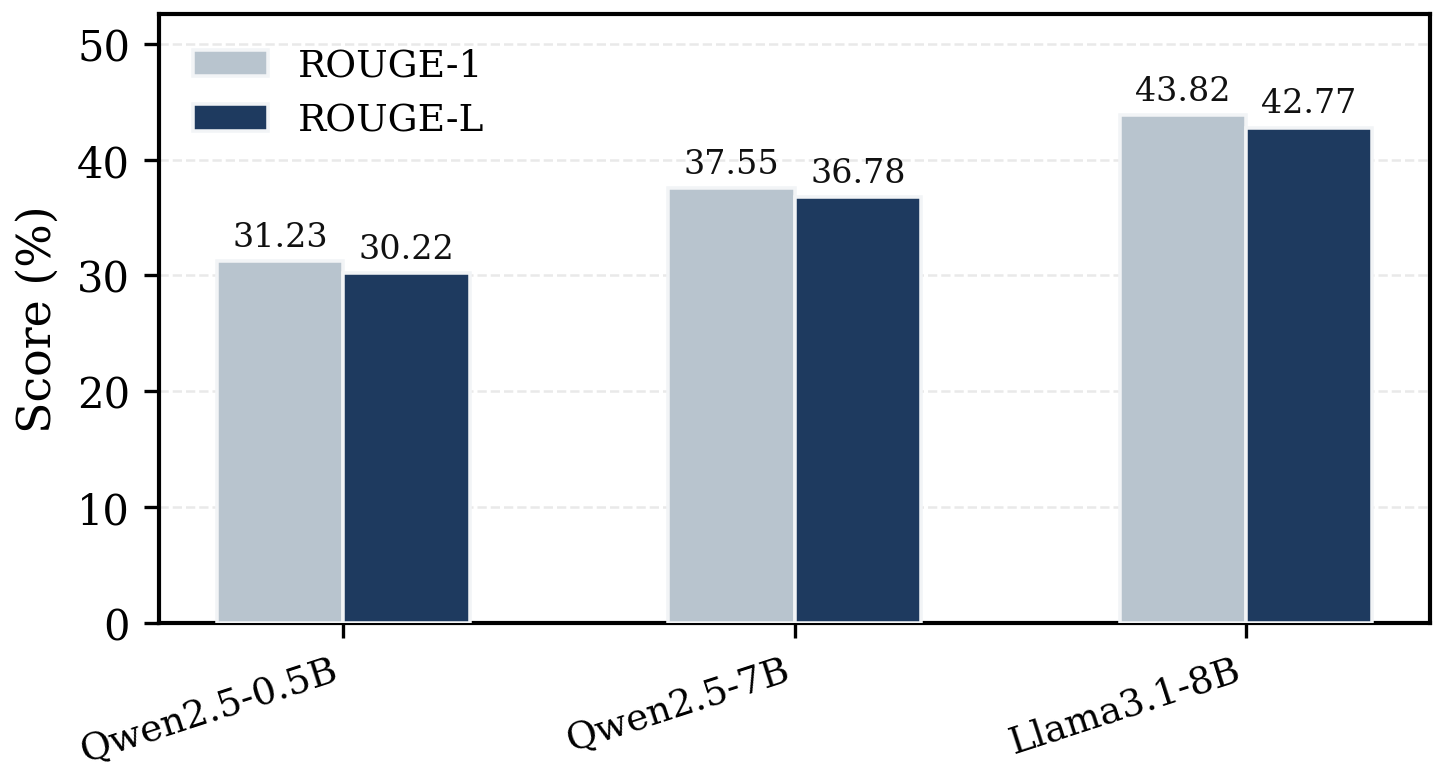}
        \vspace{-6pt}
        \caption{NarrativeQA}
        \label{fig:nqa_all_model}
    \end{subfigure}

    \begin{subfigure}{1.0\linewidth}
        \centering
        \includegraphics[width=0.9\linewidth]{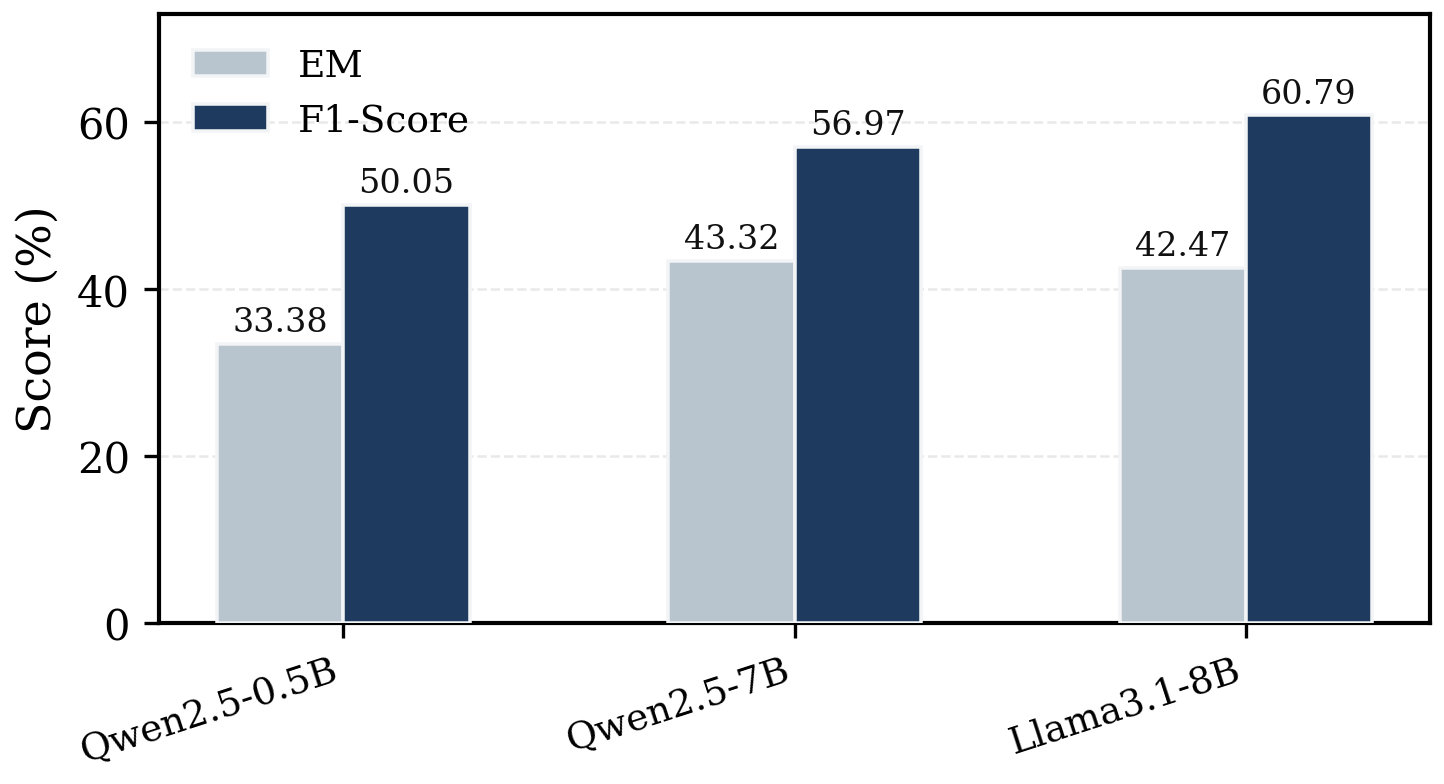}
        \vspace{-6pt}
        \caption{SQuAD}
        \label{fig:squad_all_model}
    \end{subfigure}
    \vspace{-10pt}
    \caption{Results across different models. }
    \label{fig:result_all}
    \vspace{-20pt}
\end{figure}

To further examine the robustness and scalability of our method, we evaluate it with different backbone models, including Qwen2.5-0.5B, Qwen2.5-7B, and Llama3.1-8B. The results are shown in Figure~\ref{fig:result_all}. Overall, our method consistently achieves strong performance across all evaluated backbones on both NarrativeQA and SQuAD, indicating that the proposed approach is not tied to a specific model family or parameter scale.

These results suggest that our method generalizes well across model scales and architectures. In particular, the consistent gains from smaller to larger backbones indicate that the proposed design can leverage improved model capacity, while still remaining effective even when applied to relatively compact models.

\end{document}